\documentclass[10pt,twocolumn,twoside]{IEEEtran}
\usepackage{graphicx}
\usepackage{amsmath, amssymb}
\usepackage{algorithm}
\usepackage{algorithmic}
\usepackage{array}
\usepackage{multirow}
\usepackage{subfigure}
\usepackage{picins}
\usepackage{balance}
\ifCLASSINFOpdf
\else
\fi

\begin{document}
\title{Learning Expressionlets via Universal Manifold Model for Dynamic Facial Expression Recognition}
%
%
%

\author{Mengyi Liu,~\IEEEmembership{Student Member,~IEEE},
        Shiguang Shan,~\IEEEmembership{Senior Member,~IEEE},
        Ruiping Wang,~\IEEEmembership{Member,~IEEE},\\
        Xilin Chen,~\IEEEmembership{Senior Member,~IEEE}
}

\maketitle

\begin{abstract}
  Facial expression is temporally dynamic event which can be decomposed into a set of muscle motions occurring in different facial regions over various time intervals. For dynamic expression recognition, two key issues, temporal alignment and semantics-aware dynamic representation, must be taken into account. In this paper, we attempt to solve both problems via manifold modeling of videos based on a novel mid-level representation, i.e. \textbf{expressionlet}. Specifically, our method contains three key stages: 1) each expression video clip is characterized as a spatial-temporal manifold (STM) formed by dense low-level features; 2) a Universal Manifold Model (UMM) is learned over all low-level features and represented as a set of local modes to statistically unify all the STMs. 3) the local modes on each STM can be instantiated by fitting to UMM, and the corresponding expressionlet is constructed by modeling the variations in each local mode. With above strategy, expression videos are naturally aligned both spatially and temporally. To enhance the discriminative power, the expressionlet-based STM representation is further processed with discriminant embedding. Our method is evaluated on four public expression databases, CK+, MMI, Oulu-CASIA, and FERA. In all cases, our method outperforms the known state-of-the-art by a large margin.
\end{abstract}

\IEEEpeerreviewmaketitle
\footnotetext{
M. Liu, S. Shan, R. Wang, and X. Chen are with the Key Laboratory of Intelligent Information Processing of Chinese Academy of Sciences (CAS), Institute of Computing Technology, CAS, Beijing 100190, China. (e-mail: mengyi.liu@vipl.ict.ac.cn; \{sgshan, wangruiping, xlchen\}@ict.ac.cn).

Shiguang Shan is the corresponding author of this paper.
}
\section{Introduction}
Automatic facial expression recognition plays an important role in various applications, such as Human-Computer Interaction (HCI) and diagnosing mental disorders. Early research mostly focused on expression analysis from static facial images \cite{pantic2000automatic}. However, as facial expression can be better described as the sequential variation in a dynamic process, recognizing facial expression from video is more natural and proved to be more effective in recent research works \cite{zhao2007dynamic,yang2008facial,chew2011person,zhao2011facial,sanin2013spatio}.

\begin{figure}[tbh]
\centering
\includegraphics[height=5cm]{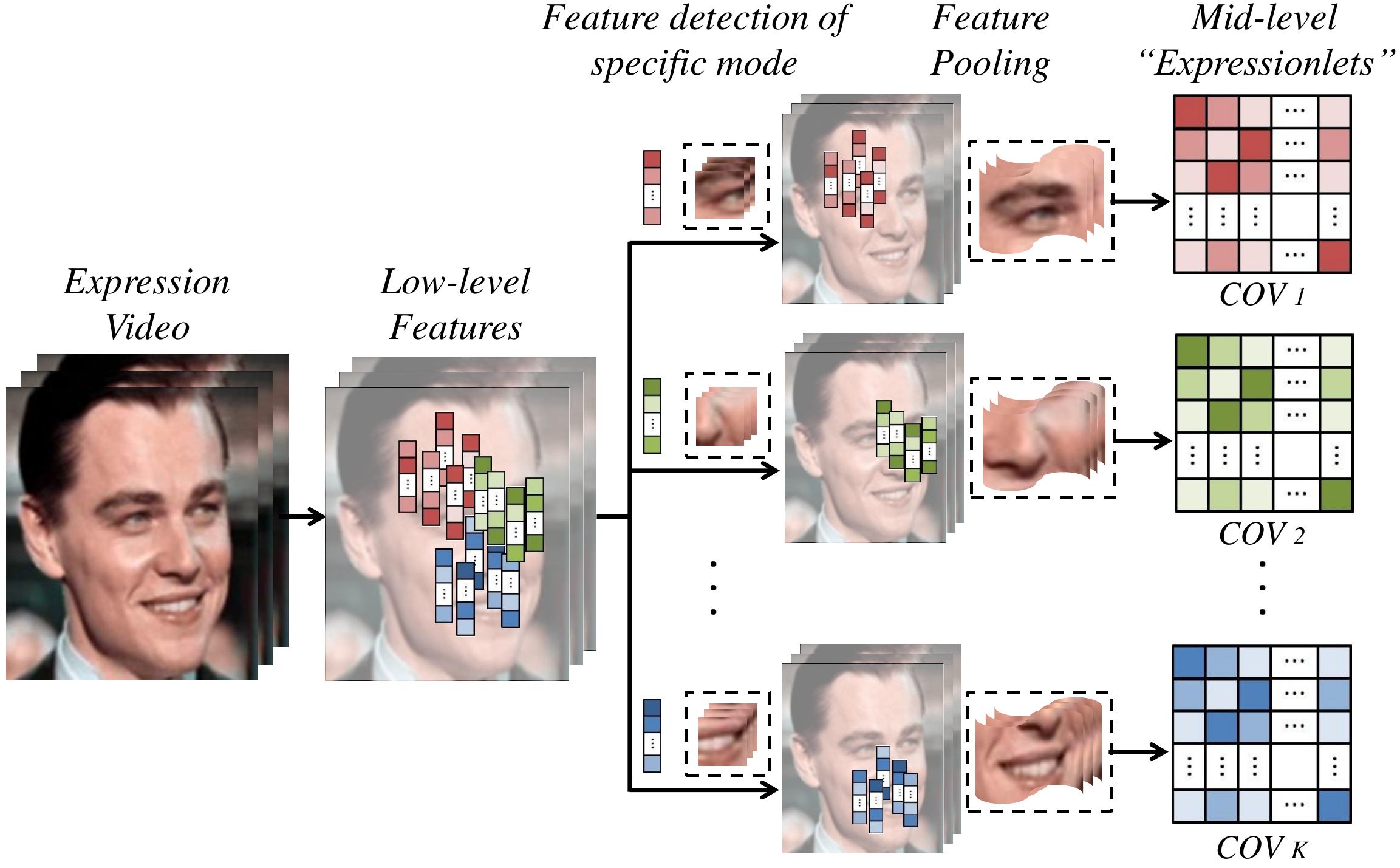}
\caption{A schematic illustration of constructing the mid-level representation -- the proposed ``expressionlets'' (``COV'' is short for ``covariance matrix''). Each strip stands for a local feature, and the K feature modes (similar to codewords) are pre-learned and modeled via GMM.}
\label{fig:figExpLet}
\vspace{-10pt}
\end{figure}

\begin{figure*}[t]
\centering
\includegraphics[height=8.5cm]{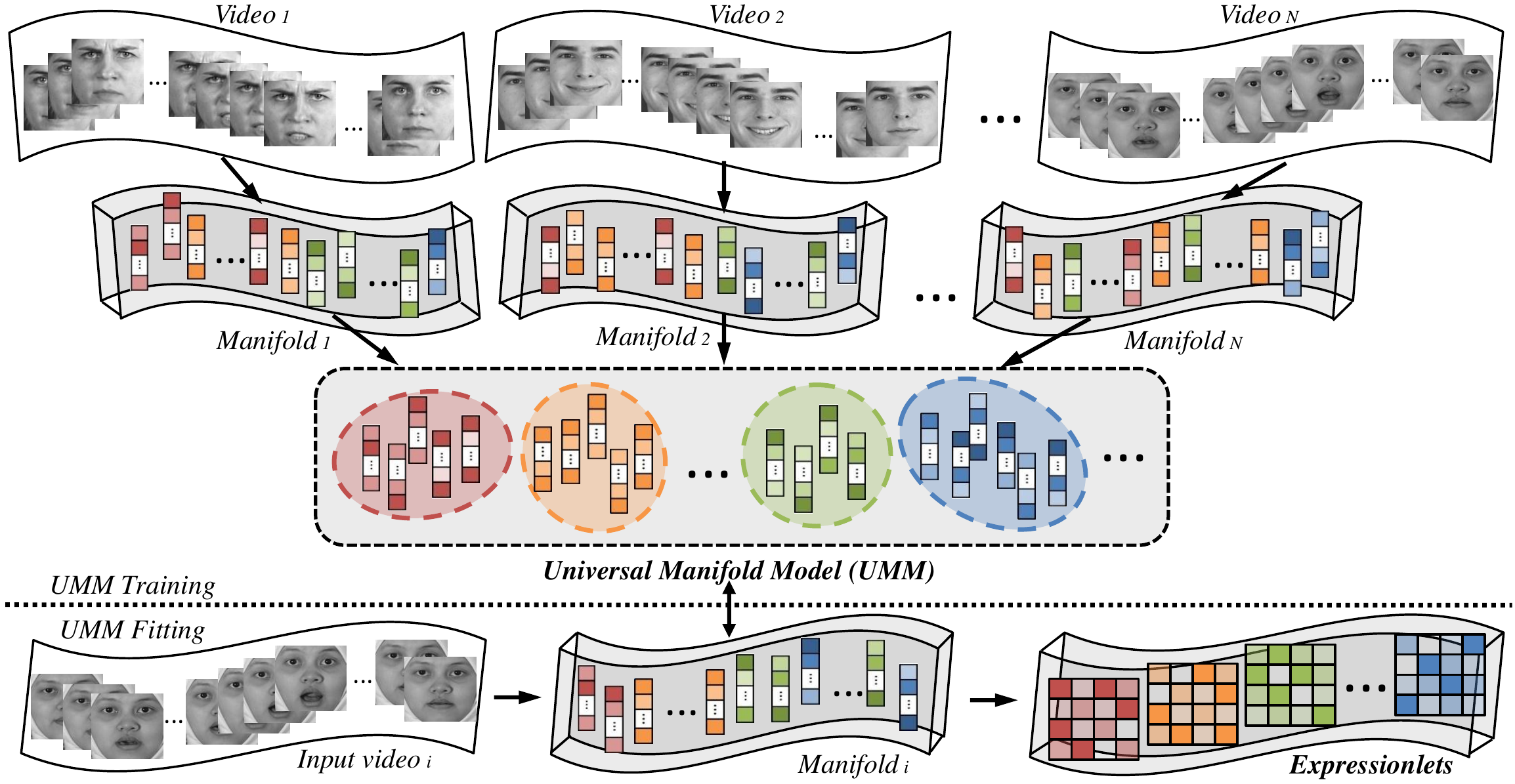}
\caption{The schema of the proposed method. Given an individual video clip, we intend to model it as a Spatial-Temporal Manifold (STM) spanned by local spatial-temporal features, which however leads to difficulty of aligning different STMs. To statistically unify and thus facilitate the alignment of STMs, we propose a Universal Manifold Model (UMM), represented as a number of universal local ST modes, which can be learned by EM-like methods. With UMM constructed, the local modes on each STM can be instantiated by fitting to UMM and thus aligned mutually, then the corresponding expressionlet is built to model the variations (via covariance matrix) in each local ST mode. Thus we obtain an expressionlet-based representation of STM. Please note that, for UMM training, we exploit both appearance and spatial-temporal location information of the local features in order to enforce some degree of locality both spatially and temporally.}
\label{fig:figUMM}
\end{figure*}

Among these video-based facial expression recognition methods, one of the main concerns is how to effectively encode the dynamic information in videos. Currently, the mainstream approaches to dynamic representation are based on local spatial-temporal features like LBP-TOP \cite{zhao2007dynamic}, HOG 3D \cite{klaser2008spatio}. These local descriptors extracted in local cuboid are then pooled over the whole video or some hand-crafted segments, to obtain a representation with certain length independent of time resolution. As the low-level features possess the property of repeatability, integrating them by pooling leads to robustness to intra-class variations and deformations of different expression styles. However, this kind of technique lacks of consideration of two important issues: 1) \textbf{Temporal alignment}. Expressions are inherently dynamic events consisting of onset, apex, and offset phases. Intuitively, the recognition should conduct matching among corresponding phases, which thus requires globally temporal alignment among different sequences. The rigid pooling has inevitably dropped those sequential relations and temporal correspondences. 2) \textbf{Semantics-aware dynamic representation}. Each expression can be decomposed into a group of semantic action units, which exhibit in different facial regions with varying sizes and last for different lengths of time. Since the manually designed cuboids can only capture low-level information short of representative and discriminative ability, they are incapable of modeling the expression dynamic in higher semantic level.

In this paper, we attempt to address both issues via spatial-temporal manifold modeling based on a set of mid-level representations, i.e. \textbf{expressionlets}. The proposed mid-level expressionlet is a kind of modeling that aims to characterize the variations among a group of low-level features as shown in Figure~\ref{fig:figExpLet}. The notation ``-let'' means that it serves as a local (both spatially and temporally) dynamic component within a whole expression process, which shares similar spirit with ``motionlet'' \cite{wang2013motionlets} in action recognition community. Thus expressionlet bridges the gap between low-level features and high-level semantics desirably. Specifically, given an individual video clip, we first characterize it as a Spatial-Temporal Manifold (STM) spanned by its low-level features. To conduct spatial-temporal alignment among STMs, we build a Universal Manifold Model (UMM), and represent it by a number of universal local ST modes, which can be learned by EM-like methods among all collection of low-level features. By fitting to UMM, the local modes on each STM can be instantiated respectively and all of the different STMs are inherently and mutually well-aligned to UMM via these corresponding modes. Finally, our expressionlet is constructed by modeling each local mode on STMs. To capture and characterize the correlations and variations among low-level features within each mode, the expressionlet comes in the form of covariance matrix of the feature set in a statistical manner, which also makes it robust to local misalignment \cite{tuzel2008pedestrian,hong2009sigma,wang2012covariance}.

To further enhance the discriminative ability of expressionlet, we perform a discriminant learning with these mid-level representations on all of the STMs. By considering the ``margin'' among corresponding expressionlets, we exploit a graph-embedding \cite{yan2007graph,wang2009manifold} method by constructing partially connected graphs to keep the links between expressionlets with the same semantics. In the end, the embedded features are correspondingly concatenated into a long vector as the final manifold (video) representation for classification. Hence, the proposed expressionlet has the following characteristics: 1) \textbf{Flexible spatial-temporal range}. i.e. varying sizes of spatial regions and temporal durations. 2) \textbf{Variation modeling}. It encodes the local variations caused by expression using a covariance matrix. 3) \textbf{Discriminative ability}. It is descriptive and contains category information for recognition.

Preliminary results of the method have been published in \cite{liu2014learning}. Compared with the conference version, this paper has made three major extensions. First, we generalize the framework to be compatible for various low-level 2D/3D descriptors to construct mid-level expressionlet. Second, we provide a more detailed comparison and discussion regarding different strategies for UMM learning, including the alignment manners of local modes in UMM training stage and the low-level feature assignment manners in UMM fitting stage. Third, more extensive experiments are carried out to evaluate each component in the method and compare with other state-of-the-art algorithms.

The rest of the paper is organized as follows: Section II briefly reviews the previous related work for dynamic facial expression recognition. Section III introduces the Universal Manifold Model, i.e. a statistical model for spatial-temporal alignment among different expression manifolds (videos). Section IV presents the mid-level expressionlet learning based on UMM and conducts detailed discussions with other related works. In Section V, we provide comprehensive evaluations of the whole framework as well as each of the building block. Experiments are conducted on four public expression databases and extensively compared with the state-of-the-art methods. Finally, we conclude the work and discuss possible future efforts in Section VI.

\section{Related works}

In the past several decades, facial expression recognition based on static images had aroused lots of interests among researchers. For facial feature representation, typical image descriptors including Local Binary Pattern (LBP) \cite{shan2009facial}, Local Gabor Binary Pattern (LGBP) \cite{sun2008facial}, Histogram of Oriented Gradient (HOG) \cite{hu2008multiview}, and Scale Invariant Feature Transform (SIFT) \cite{tariq2011emotion} have been successfully applied in this domain. Lucey et al. \cite{lucey2007investigating} also applied Active Appearance Model (AAM) to encode both shape (facial landmarks) and appearance variations. A comprehensive survey of some of these techniques can be found in \cite{pantic2000automatic} and \cite{zeng2009survey}.

However, as facial expressions are more naturally viewed as dynamic events involving facial motions over a time interval, recently, there becomes strong interest in modeling the temporal dynamics of facial expressions in video clips. The psychological experiments conducted in \cite{ambadar2005deciphering} have provided evidence that facial dynamics modeling is crucial for interpreting and discriminating facial expressions. Generally, the temporal modeling manners can be categorized into two groups: hard-coded and learning-based. In this paper, we review some related works of dynamic facial expression recognition based on the two schemes mentioned above.

The hard-coded modeling scheme encodes the variations among several successive frames using predefined computations. For example, optical flow is calculated between consecutive frames and has been applied in some early works for expression recognition \cite{yacoob1996recognizing,cohn1998feature}. Koelstra et al. \cite{koelstra2010dynamic} used Motion History Images (MHI) to compress the motions over several frames into a single image by layering the pixel differences between consecutive frames. Another kind of typical implementation is designing spatial-temporal local descriptors to capture the dynamic information. For instance, Yang et al. \cite{yang2008facial} designed dynamic binary patterns mapping for temporally clustered Haar-like features and adopted boosting classifiers for expression recognition. Zhao et al. \cite{zhao2011facial} encoded spatial-temporal information in image volumes using LBP-TOP \cite{zhao2007dynamic} and employed SVM and sparse representation classifier for recognition. Hayat et al. \cite{hayat2012evaluation} evaluated various dynamic descriptors including HOG/HOF \cite{laptev2008learning}, HOG3D \cite{klaser2008spatio}, and 3D SIFT \cite{scovanner20073} using bag of features framework for video-based facial expression recognition. All these methods benefit from the low computational cost of local descriptors and also show favourable generalizations to different data sources and recognition tasks.

To consider the specific characteristics of dynamic facial expressions, the learning-based modeling schemes attempt to explore the intrinsic correlations among facial variations using dynamic graphical models. Some representative works are briefly introduced as follows: Cohen et al. \cite{cohen2003facial} used Tree-Augmented Naive Bayes (TAN) classifier to learn the dependencies among the facial motion features extracted from a continuous video. Shang et al. \cite{shang2009nonparametric} applied a non-parametric discriminant Hidden Markov Model (HMM) on the facial features tracked with Active Shape Model (ASM) to recognize dynamic expressions. Jain et al. \cite{jain2011facial} proposed a framework by modeling temporal variations within facial shapes using Latent-Dynamic Conditional Random Fields (LDCRFs), which obtains the entire video prediction and continuously frame labels at the same time. To further characterize the complex activities both spatially and temporally, Wang et al. \cite{wang2013capturing} proposed Interval Temporal Bayesian Networks (ITBN) to represent the spatial dependencies among primary facial events and the large variety of time-constrained relations simultaneously. To summarize, the learning-based modeling can better reveal the intrinsic principles of the dynamic variations caused by facial expressions. However the construction and optimization of a such model required lots of domain knowledge and high computational cost.

\section{Universal Manifold Model (UMM)}

A facial expression video depicts continuous shape or appearance variations and can be naturally modeled by a non-linear manifold, on which each point corresponds to a certain local spatial-temporal pattern. For dynamic expression recognition, the main challenge is the large arbitrary inter-personal variance of expressing manners and execution rate for the same expression category, thus it is crucial to conduct both spatial and temporal alignment among different expression manifolds. In this section, we first introduce the manifold modeling of videos and then propose a statistic-based Universal Manifold Model (UMM) to achieve implicit alignment among different expression videos.

\subsection{Spatial-Temporal Manifold}

For clarification, we first present the spatial-temporal manifold (STM) for modeling each video clip. The STM is spanned by 3D (i.e. spatial-temporal) blocks densely sampled from the video volume, which cover a variety of local variations in both spatial and temporal space. Two kinds of common descriptors, i.e. SIFT and HOG, are employed for low-level feature extraction on each sampled block with the size of $w*h*l$, where $w,h$ are the numbers of pixels on two spatial directions, and $l$ is the number of frames. The extracted feature is denoted as $a_{xyt}$, where $x,y,t$ are spatial-temporal index of the block on the STM.

\begin{figure}[tbh]
\centering
\includegraphics[height=3.8cm]{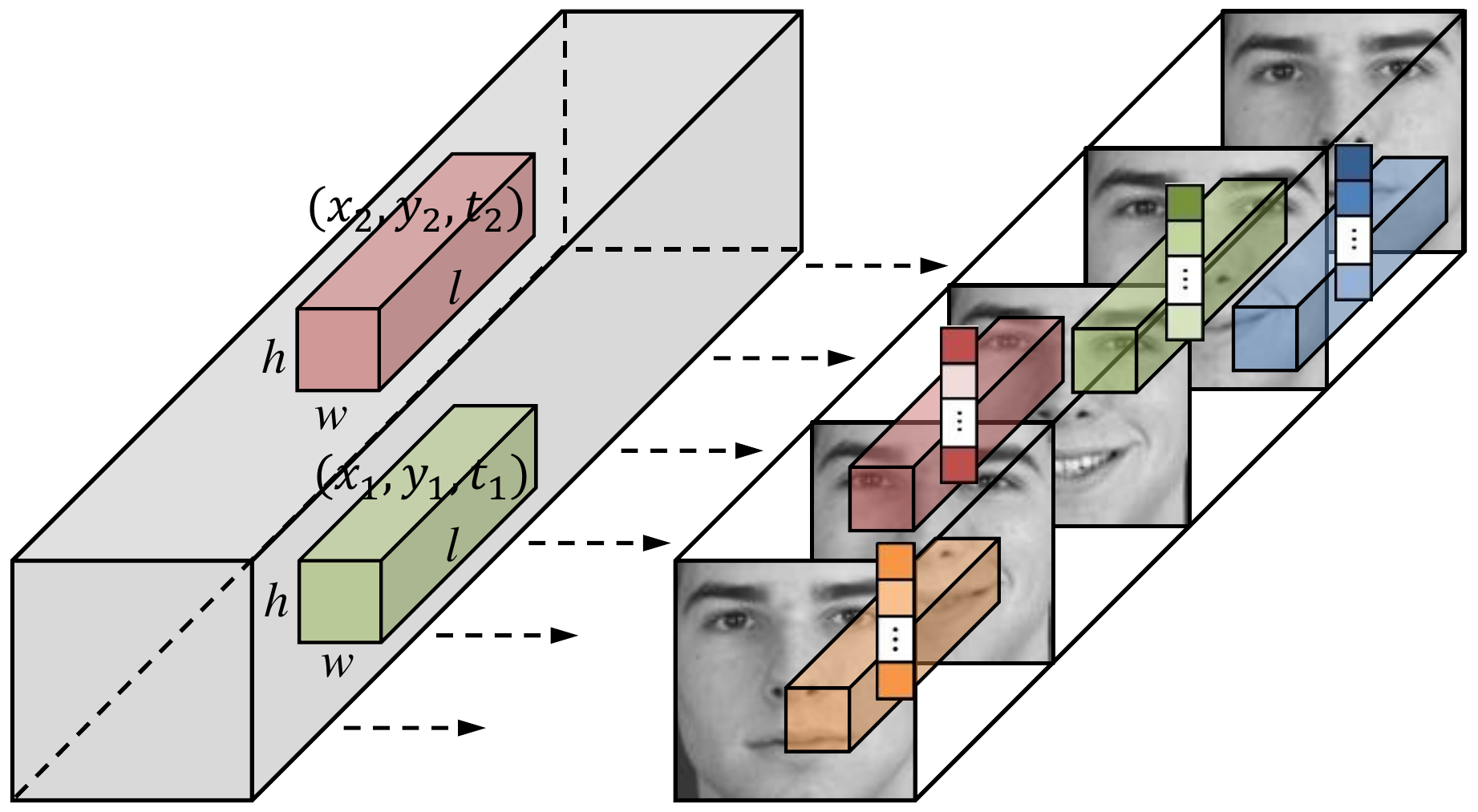}
\caption{An illustration of the spatial-temporal blocks for low-level feature extraction. The augmented feature are then used to construct the STM.}
\label{fig:figExpSTM}
\end{figure}

To consider the manifold structure information, for all the blocks we augment the appearance features with their spatial-temporal coordinates, i.e. $f=\{a_{xyt},x/w^*,y/h^*,t/l^*\}$, where $a_{xyt}$ is the appearance feature of the block located at $\{x,y,t\}$, and $w^*,h^*,l^*$ are the numbers of blocks on width, height and time length direction on the STM. An illustration of the local features is shown in Figure~\ref{fig:figExpSTM}.

\subsection{UMM Learning}

\subsubsection{Training stage} Universal Manifold Model (UMM) is defined to statistically model the STMs from different people with different expressions. As a person-independent and expression-independent model, UMM facilitates the robust parameterized modeling of the STMs. Inspired by \cite{hasan2011study,li2013probabilistic}, we employ a Gaussian Mixture Model (GMM) to learn the UMM by estimating the appearance and location distribution of all the 3D block features. Thus each Gaussian component can represent a certain spatial-temporal mode modeling the variations among a set of low-level features with similar appearance and relative locations in videos.

Simply we can train a GMM with spherical Gaussian components as follows:
\vspace{-5pt}
\begin{equation}
P(f|\Theta)=\displaystyle{\sum_{k=1}^K}w_k G(f|\mu_k,\sigma_k^2I),
\end{equation}
where $\Theta=(w_1,\mu_1,\sigma_1,...,w_K,\mu_K,\sigma_K)$; $K$ is the number of Gaussian mixture components; $I$ is identity matrix; ${w_k,\mu_k,\sigma_k}$ are the mixture weight, mean, and diagonal covariance of the $k$-th Gaussian component $G(f|\mu_k,\sigma_k)$. We use typical Expectation Maximization (EM) algorithm to estimate the paremeters of GMM by maximizing the likelihood of the training feature set. After training the UMM, each Gaussian component builds correspondence of a group of block features from different STMs, which constitute a local ST mode universally.

\subsubsection{Fitting stage} The UMM learned above can be regarded as a container with K-components GMM. Then, given any STM, we aim to formulate it as a parameterized instance of the UMM. For this purpose, our basic idea is assigning some of the local ST features of the STM into the K Gaussian "buckets" and further modeling the distribution of the local features in each Gaussian bucket with their covariance matrix.

Formally, an expression manifold $M^i$ can be presented as a set of local block features, i.e. $\mathcal{F}^i=\{f_1^i,...,f_{B_i}^i\}$, where $B_i$ is the number of features on $M^i$. For the $k$-th Gaussian component $G(f|\mu_k,\sigma_k)$ on UMM, we can calculate the probabilities of each $f_b^i$ in $F^i$ as
\begin{equation}
\begin{array}{rl}
P_k^i=\{p_k(f_b^i)~|~p_k(f_b^i)=w_kG(f_b^i|\mu_k,\sigma_k^2I)\}_{b=1}^{B_i}.\\
\end{array}
\end{equation}

\begin{figure}[tbh]
\centering
\includegraphics[height=3.2cm]{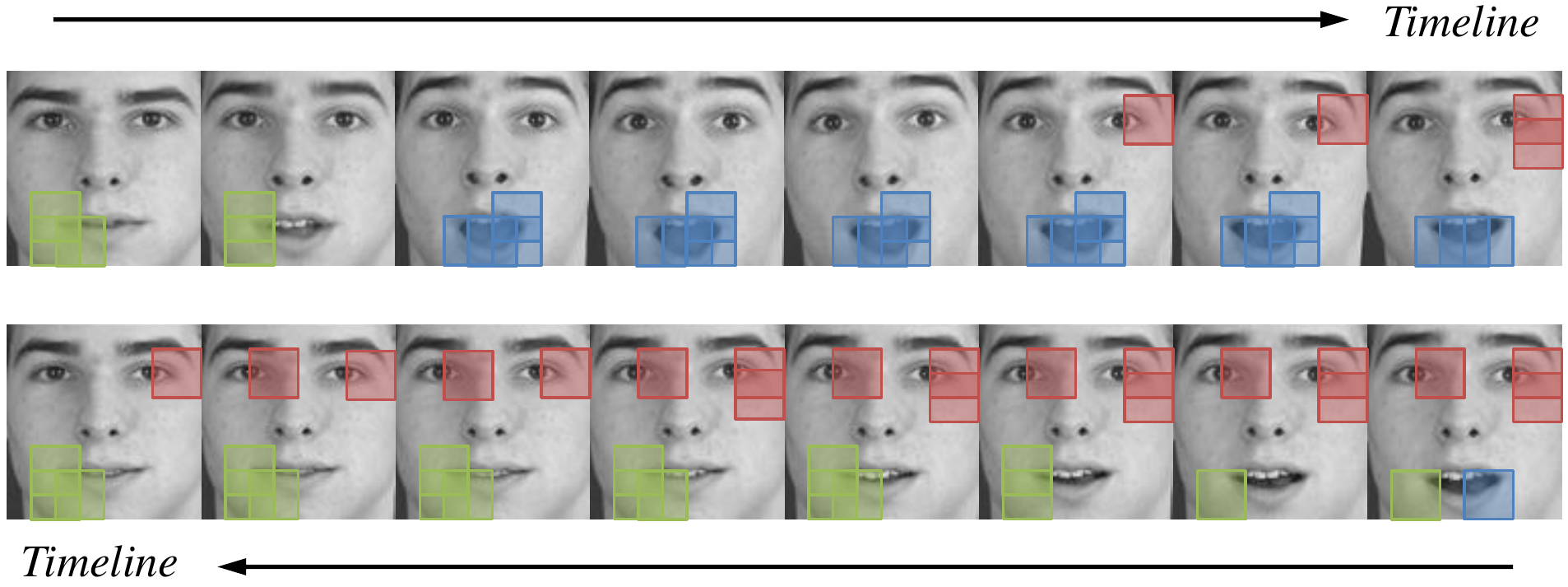}
\caption{Examples of typical local modes (a set of features with largest $T$ ($T=24$ in the examples) probabilities to a certain Gaussian component on UMM) referring to the original spatial-temporal locations in videos. Different colors represent different local modes. \textbf{Best viewed in color.}}
\label{fig:figExpUMM}
\end{figure}

We sort the block features $f_b^i$ in descending order of $P_k^i$, and the top $T$ features with the largest probabilities are selected for the $k$-th local mode construction, which can be represented as $F_k^i=\{f_{k_1}^i,...,f_{k_T}^i\}$. The selected features in each set are expected to be close in space-time location and share similar appearance characteristics, which can represent the local variations occurred in a certain facial region during a small period of time. Different from the hard assignment in traditional GMM, by using such soft manner, one feature can be assigned to multiple modes (components) for sharing, which brings favorable robustness against mis-assignment. Moreover, discarding some useless features with low probabilities to any mode can also be regarded as a ``filtering'' operation, which can alleviate the influence of unexpected noises irrelevant to expressions. In Figure~\ref{fig:figExpUMM}, we also demonstrate some examples of the learned local modes referring to the original spatial-temporal locations in videos.

Finally, an overall procedure is summarized in Algorithm~\ref{alg:algUMM}. Based on the input unaligned STMs $\mathcal{F}^1,...,\mathcal{F}^N$, each of which is represented by a set of low-level features, the algorithm provides two kinds of outputs: a group of learned optimal UMM parameters $\Theta^*$, and the mutually aligned STMs $\mathcal{\widetilde{F}}^1,...,\mathcal{\widetilde{F}}^N$, each of which is represented by $K$ corresponding local modes instantiated by fitting to UMM.

\begin{algorithm}[tbh]
\caption{\textbf{:~UMM Learning}}
\renewcommand{\algorithmicrequire}{\textbf{Input:}}
\renewcommand\algorithmicensure {\textbf{Output:}}
\begin{algorithmic}[1]
\label{alg:algUMM}
\vspace{3pt}
\REQUIRE ~~\\
Unaligned STMs (represented by sets of low-level features):~$\mathcal{F}^1,...,\mathcal{F}^N$\\
\vspace{3pt}
\ENSURE ~~\\
Mutually aligned STMs (represented by corresponding local modes instantiated by fitting to UMM):~$\mathcal{\widetilde{F}}^1,...,\mathcal{\widetilde{F}}^N$ \\
\vspace{3pt}
{\textbf{--- Training ---}}
\vspace{3pt}
\STATE Initialize UMM (GMM) parameter:~$\Theta=\{(\omega_k,\mu_k,\sigma_k)\}$
\vspace{3pt}
\STATE Use EM algorithm to learn optimal UMM parameters:\\
\vspace{3pt}
$\Theta^*=\mathop{argmax}_\Theta \sum_{i,b,k}\omega_kG(f_b^i|\mu_k,\sigma_k^2I)$ \\
\vspace{3pt}
{\textbf{--- Fitting ---}}
\vspace{3pt}
\FOR{i:=1 to N}
\vspace{3pt}
\FOR{k:=1 to K}
\vspace{3pt}
\STATE Find top $T$ block features $F^i_k=\{f_{k_t}^i\}_{t=1}^T$ with the largest probabilities on $G_k$: \\
\vspace{3pt}
$G(f_{k_t}^i|\mu_k^*,(\sigma_k^*)^2I)>G(f_{k_{t+1}}^i|\mu_k^*,(\sigma_k^*)^2I)$ \\
\vspace{3pt}
\ENDFOR
\vspace{3pt}
\STATE $\mathcal{\widetilde{F}}^i=\{F_1^i,F_2^i,...,F_K^i\}$
\vspace{3pt}
\ENDFOR
\vspace{3pt}
\RETURN $\Theta^*$, $\mathcal{\widetilde{F}}^1,...,\mathcal{\widetilde{F}}^N$
\end{algorithmic}
\end{algorithm}

\section{Expressionlet learning}

\subsection{Expressionlet modeling}

Considering the correlations and variations among the features in a local model, we calculate the covariance matrix of the set $F_k^i$ as the representation of an expressionlet:
\begin{equation}
C_k^i=\frac{1}{T-1}\displaystyle{\sum_{t=1}^T}(f_{k_t}^i-\overline{f_k^i})(f_{k_t}^i-\overline{f_k^i})^T,
\end{equation}
where $\overline{f_k^i}$ is the mean of the block features in set $F_k^i$. The diagonal entries of $C_k^i$ represent the variance of each individual feature, and the non-diagonal entries are their respective correlations. As the expressionlets are globally aligned via UMM, the covariance modeling can provide a desirable locally tolerance to spatial-temporal misalignment.

In the end, the $i$-th manifold $M^i$ can be represented as a set of expressionlets, i.e. $E^i=\{C_1^i,C_2^i,...,C_K^i\}$. Here the expressionlets are Symmetric Positive Definite (SPD) matrices (i.e. nonsingular covariance matrices), lying on a Riemannian manifold \cite{pennec2006riemannian}. We exploit a Log-Euclidean Distance (LED) \cite{arsigny2007geometric} to project these points to Euclidean vector space, where standard vector learning methods are ripely studied, as advocated in \cite{wang2012covariance}.

Given a covariance matrix $C$, the mapping to vector space is equivalent to embedding the SPD manifold $\mathcal{M}$ into its tangent space $\mathcal{T}$ at identity matrix $I$, i.e.:
\begin{equation}
\Psi: \mathcal{M} \mapsto \mathcal{T}_I, C \mapsto (log(C)).
\label{equ:equLog}
\end{equation}

Let $C=U\Sigma U^T$ be the eigen-decomposition of SPD matrix $C$, its $log$ can be computed by
\begin{equation}
log(C)=Ulog(\Sigma)U^T.
\end{equation}

As we obtain a vector mapping of $C$ spanned by $log(C)$, general vector learning methods, e.g. PCA, can be employed to reduce the high dimension of expressionlet. Basically, in this work, we preserve 99\% energies for the expressionlets using PCA for further discriminant learning.

\subsection{Discriminant learning with Expressionlets}
\label{sec:secDisExpLet}

As the expressionlet possesses the property of spatial-temporal locality, an effective way of enhancing its discriminative power is to consider the ``margin'' among corresponding expressionlets from different STM samples. Thus we can formulate our learning scheme via the graph embedding \cite{yan2007graph} framework.

\begin{figure}[tbh]
\centering
\includegraphics[height=3.2cm]{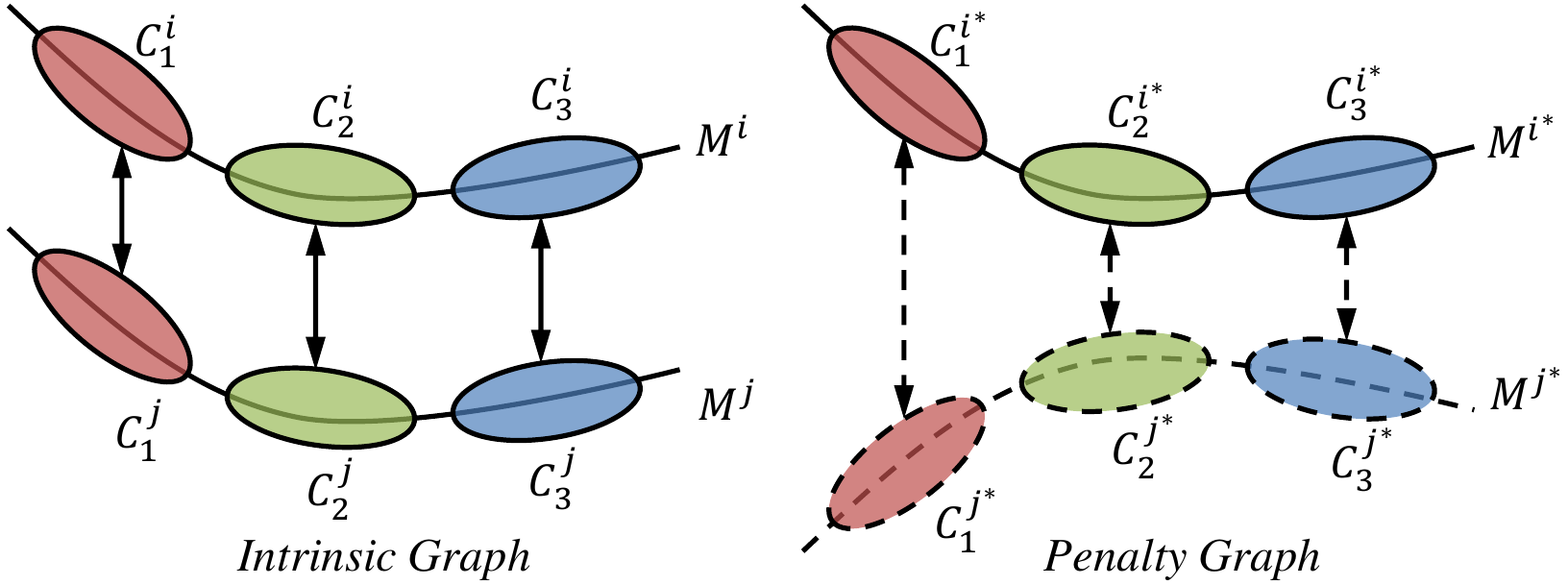}
\caption{The adjacency relationships of the intrinsic and penalty graphs for the discriminative learning with expressionlets (Different colors represent the different Gaussian components in UMM). $M^i$ and $M^j$ are two manifolds with the same class label, while $M^{i^*}$ and $M^{j^*}$ are with different class labels. The intrisic/penalty graph only considers the ``margin'' among corresponding expressionlets ($C_k^i$ and $C_k^j$) generated from the same Gaussian component $k$.}
\label{fig:figDiscriminant}
\end{figure}

In the overall expressionlet set $\{E^1,...,E^N\}$, given the $m$-th expressionlet, which corresponds to the $p$-th mode on $M^i$, denoted as $C_p^i$; and the $n$-th expressionlet, which corresponds to  the $q$-th mode on $M^j$, denoted as $C_q^j$ (Note that, if all STMs are ordered, we can denote $m=(i-1)*K+p$ and similarly $n=(j-1)*K+q$. The indices $m$ and $n$ are used for better illustration), with the class label $l_i, l_j$ for $M_i, M_j$ respectively, the intrinsic graph $W_w$ and penalty graph $W_b$ can be defined as follows:
\begin{equation}
W_w(m,n)=\left\{
\begin{aligned}
&1,~if~l_i=l_j, and~p=q \\
&0,~otherwise
\end{aligned}
\right.
\end{equation}
\begin{equation}
W_b(m,n)=\left\{
\begin{aligned}
&1,~if~l_i\neq l_j, and~p=q \\
&0,~otherwise
\end{aligned}
\right.
\end{equation}

We aim to learn an embedding function $\phi$ to maximize the discriminative power while simultaneously preserve the correspondence of expressionlets from the same Gaussian component. According to $W_w$ and $W_b$, the within-class scatter $S_w$ and between-class scatter $S_b$ can be defined as:
\begin{equation}
S_w=\sum_{m,n}Dis(\phi(C_p^i),\phi(C_q^j))W_w(m,n),
\end{equation}
\begin{equation}
S_b=\sum_{m,n}Dis(\phi(C_p^i),\phi(C_q^j))W_b(m,n),
\end{equation}
where $Dis(\phi(C_p^i),\phi(C_q^j))$ denotes the distance between two embedded expressionlets $\phi(C_p^i)$ and $\phi(C_q^j)$.

According to Equation~\ref{equ:equLog} we can obtain a vector representation $x_m$ of the $m$-th expressionlet, i.e. $C_p^i$, where $x_m$ is a vector spanned by $log(C_p^i)$. Simply consider a linear projection $v$, we can reformulate the embedded features and the distance between them in classical Euclidean space as
\begin{equation}
\phi(C_p^i)=v^Tx_m, \phi(C_q^j)=v^Tx_n,
\end{equation}
\begin{equation}
Dis(\phi(C_p^i),\phi(C_q^j))=||v^Tx_m-v^Tx_n||^2.
\end{equation}

Accordingly, we only need to learn the projection $v$ instead of $\phi$, by maximizing the between-class scatter $S_b$ while minimizing the within-class scatter $S_w$:
\begin{equation}
v_{opt}=\arg\max\frac{v^TX(D_b-W_b)X^Tv}{v^TX(D_w-W_w)X^Tv},
\end{equation}
where $D_w$ and $D_b$ are diagonal matrices with diagonal elements $D_w(m,m)=\sum_nW_w(m,n)$ and $D_b(m,m)=\sum_nW_b(m,n)$. Let $L_w$ and $L_b$ be the Laplacian matrices of two graphs $W_w$ and $W_b$. The columns of an optimal $v$ are the generalized eigenvectors corresponding to the $l$ largest eigenvalues in
\begin{equation}
XL_bX^Tv=\lambda XL_wX^Tv.
\end{equation}

With the learned embedding function $\phi$, the $K$ expressionlets from $M_i$ can be represented as $\{\phi(C_1^i),...,\phi(C_K^i)\}$. These $K$ features are concatenated to form a long vector as the final expression manifold (video) representation. In the end, we use multi-class linear SVM implemented by Liblinear \cite{fan2008liblinear} for classification.

\subsection{Discussion}

\subsubsection{Expressionlet vs. AU}

Action Units (AU) \cite{ekman1978facial} are fundamental actions of individual or groups of facial muscles for encoding facial expression based on Facial Action Coding System (FACS). Similarly, our expressionlets are designed to model expression variations over local spatio-temporal regions in the same spirit as AUs. However, there are two differences between expressionlets and AUs: (i) AUs are manually defined concepts that are independent of person and category, while expressionlets are some mid-level representations extracted from data using learning scheme, which possess the dynamic modeling ability and discriminative power. (ii) According to FACS, each expression is encoded by the existence of a certain number of AUs. Instead of the binary coding manner, in our method, an expression can be represented by various real-valued expressionlet patterns which provide more flexible and rich information.

\subsubsection{Expressionlet vs. BoVW/VLAD/FV}

In our method, we extract dense local spatial-temporal features and construct a codebook (via GMM), in which each codeword can be considered as a representative of several similar local features. Both of the two operations (i.e. local feature extraction, and codebook construction) are also typical steps in Bag of Visual Words (BoVW) (or Vector of Locally Aggregated Descriptors (VLAD), and Fisher Vectors (FV)) framework.

However, in pooling stage, BoVW/VLAD/FV all perform summing/accumulating operation among the local features assigned to each certain codeword. Specifically, BoVW \cite{csurka2004visual} simply estimates histogram(s) of occurrences of each codeword; VLAD accumulates the first-order difference of the vectors assigned to each codeword, which characterizes the distribution with respect to the center (codeword) \cite{jegou2010aggregating}; Compared to VLAD, FV encodes both first-order and second-order statistics of the difference between the codewords and pooled local features and accumulates them based on the Gaussian component weights of GMM learned for codebook construction \cite{perronnin2010improving}. However, in our method, different from the summing operation, we make use of the second-order statistics by estimating the covariance of all the local features (augmented with location information) falling into each bucket (codeword). In this way, the local features are pooled to keep more variations, which not only encodes the relationship (difference) between the center and pooled features, but also includes the internal correlations among those pooled features which collaboratively describe some kind of motion patterns (i.e. expressionlets). In addition, in our method, by limiting the number ($T$ in Algorithm~\ref{alg:algUMM}) of local features falling into each bucket, not all local features are necessarily taken into account by the second-order pooling, which is also different from traditional methods. We believe such a strategy can alleviate the influence of unexpected noise or signal distortions (e.g. caused by occlusion).

\section{Experiments}

\subsection{Datasets and protocols}

\subsubsection{CK+ database} The CK+ database \cite{lucey2010extended} consists of 593 sequences from 123 subjects, which is an extended version of Cohn-Kanade (CK) database. The image sequence vary in duration from 10 to 60 frames and incorporate the onset (neutral face) to peak formation of the facial expression. The validated expression labels are only assigned to 327 sequences which are found to meet the criteria for 1 of 7 discrete emotions (Anger, Contempt, Disgust, Fear, Happiness, Sadness, and Surprise) based on Facial Action Coding System (FACS). We adopt leave-one-subject-out cross-validation (118 folds) following the general setup in \cite{lucey2010extended}.

\subsubsection{Oulu-CASIA database} 
\begin{figure}[tbh]
\centering
\includegraphics[height=4.5cm]{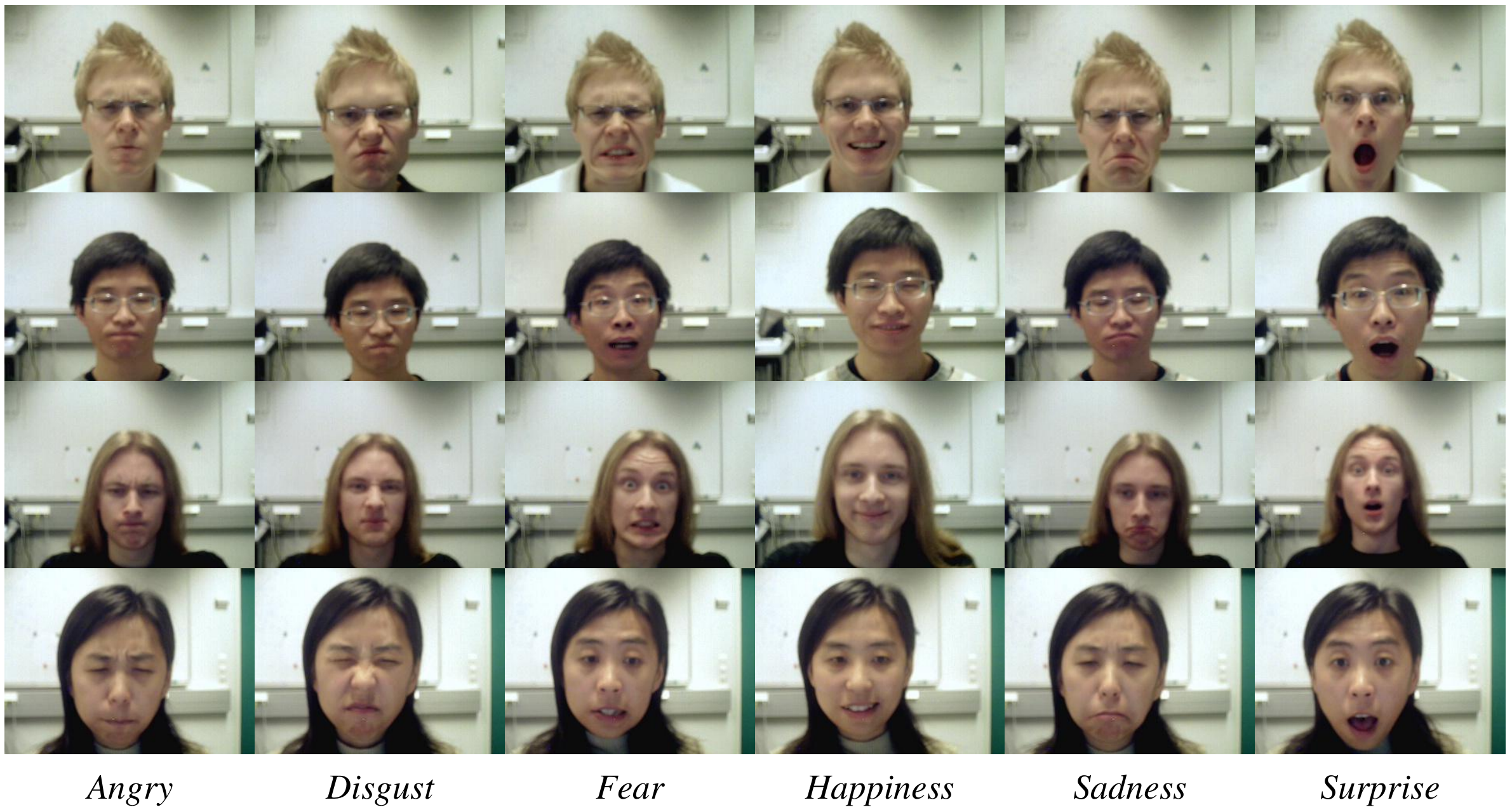}
\caption{The sample facial expression images extracted from the apex frames of video from Oulu-CASIA databse.}
\label{fig:figImageOulu}
\end{figure}

The Oulu-CASIA VIS database \cite{zhao2011facial} is a subset of the Oulu-CASIA NIR-VIS database, in which all the videos were taken under the visible (VIS) light condition. We evaluated our method only on the normal illumination condition (i.e. strong and good lighting). It includes 80 subjects between 23 and 58 years old, with six basic expressions (i.e. anger, disgust, fear, happiness, sadness, and surprise) of each person. Each video starts at a neutral face and ends at the apex of expression as the same settings in CK+. Similar to \cite{zhao2011facial} and \cite{guo2012dynamic}, we adopted person-independent 10-fold cross-validation scheme on the total 480 sequences. Figure~\ref{fig:figImageOulu} shows some sample facial expression images extracted from the apex frames of video from Oulu-CASIA databse.

\subsubsection{MMI database} The MMI database \cite{valstar2010induced} includes 30 subjects of both sexes and ages from 19 to 62. In the database, 213 sequences have been labeled with six basic expressions, in which 205 sequences were captured in frontal view. Each of the sequence reflects the whole temporal activation patterns (onset $\rightarrow$ apex $\rightarrow$ offset) of a single facial expression type. In our experiments, all of these data were used and also a person-independent 10-fold cross-validation was conducted as in several previous work \cite{guo2012dynamic,liu2014learning}. Compared with CK+ and Oulu-CASIA, MMI is thought to be more challenging for the subjects pose expressions non-uniformly and usually wear some accessories (e.g. glasses, moustache). The number of video samples for each expression in the three databases is illustrated in Table~\ref{tab:tabNumSample}.

\begin{table}[tbh]
\renewcommand\arraystretch{1.5}
\caption{The number of samples for each expression in CK+, Oulu-CASIA, and MMI databases.}
\centering
\begin{tabular}{c|ccccccc|c}
  \hline\hline
  & An & Co & Di & Fe & Ha & Sa & Su & ~~Total~~ \\
  \hline
  ~~CK+~~ & 45 & 18 & 59 & 25 & 69 & 28 & 83 & 327 \\
  ~~Oulu~~ & 80 & -- & 80 & 80 & 80 & 80 & 80 & 480 \\
  ~~MMI~~ & 31 & -- & 32 & 28 & 42 & 32 & 40 & 205 \\
  \hline\hline
\end{tabular}
\label{tab:tabNumSample}
\end{table}

\subsubsection{FERA database} 
\begin{figure}[tbh]
\centering
\includegraphics[height=3.6cm]{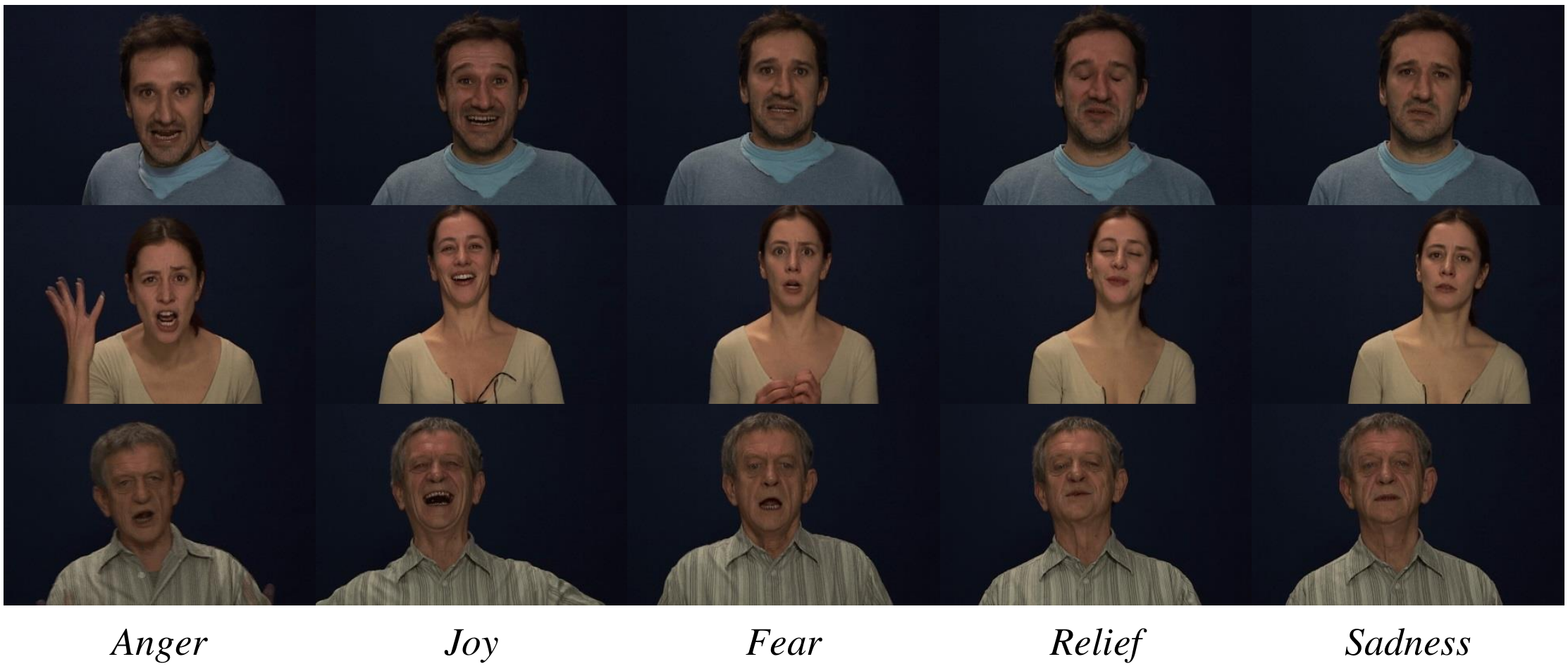}
\caption{The sample facial expression images extracted from the key frames of video from FERA databse.}
\label{fig:figImageFera}
\end{figure}

The FERA database \cite{valstar2012meta} is a fraction of the GEMEP corpus \cite{banziger2010introducing} that has been put together to meet the criteria for a challenge on facial AUs and emotion recognition. For the emotion sub-challenge, a total of 289 portrayals were selected: 155 for training and 134 for testing. The training set included 7 (3 men) actors with 3 to 5 instances of each emotion per actor, and the test set includes 6 actors, each of whom contributed 3 to 10 instances per emotion. As the labels on test set remain unreleased, we only use the training set and adopt leave-one-subject-out cross-validation for evaluation. The 155 sequences in training set have been labeled with 5 expression categories: Anger (An), Fear (Fe), Joy (Jo), Sadness (Sa), and Relief (Re). FERA is more challenging than CK+, Oulu and MMI because the expressions are \textbf{spontaneous} in natural environment. Figure~\ref{fig:figImageFera} shows some sample facial expression images extracted from the apex frames of video from FERA databse.

\subsection{Parameter settings}

For preprocessing, all the faces images are normalized to 96x96 pixels based on the locations of two eyes. In the STM construction step, the low-level 3D blocks are $w*h*l$ pixels and sampled with a stride of $0.5*w$ in spatial dimension and one frame in temporal dimension. Here $w,h$ are tunable parameters varying in 16,24,32 (the evaluations are provided in the next subsection). Two kinds of descriptors, SIFT and HOG, are employed for low-level feature extraction. For SIFT, we apply the descriptor to the center point of each $w*h*1$ patch and obtain a typical $4*4*8=128$ dimensions feature vector. PCA is further applied to reduce the dimension to 64. For HOG, each $w*h*4$ block is divided into $2*2*2$ grids and in each grid, the gradient orientations are quantized to 8 histogram bins, thus results in $2*2*2*8=64$ dimensions for each block.

In the following, we conduct detailed discussions on each framework component: (i) The effect of spatial scale for low-level feature extraction, which involves the parameter of patch size $w,h$; (ii) The effect of alignment via UMM. We compare the rigid blocking and elastic alignment manners for $K$ local modes construction, which involves the parameter of number of modes (i.e. Gaussian components in UMM); (iii) The effect of low-level feature assignment manner in UMM fitting. Both hard-assignment and soft-assignment manners are compared and discussed regarding to the parameter of number of low-level features $T$ to construct an expressionlet; (iv) The effect of discriminant learning with expressionlets. The high-dimensions of expressionlets can be reduced simply by unsupervised PCA in vector space, or a marginal discriminant learning introduced in Section~\ref{sec:secDisExpLet}. The performance of these two schemes are compared and discussed regarding to the parameter of reduced dimension $dim$ for an expressionlet.

\subsection{Evaluations of framework components}

\subsubsection{The effect of spatial scale for low-level feature extraction}

\begin{figure}[tbh]
\centering
\subfigure[CK+]{
\includegraphics[height=3.2cm]{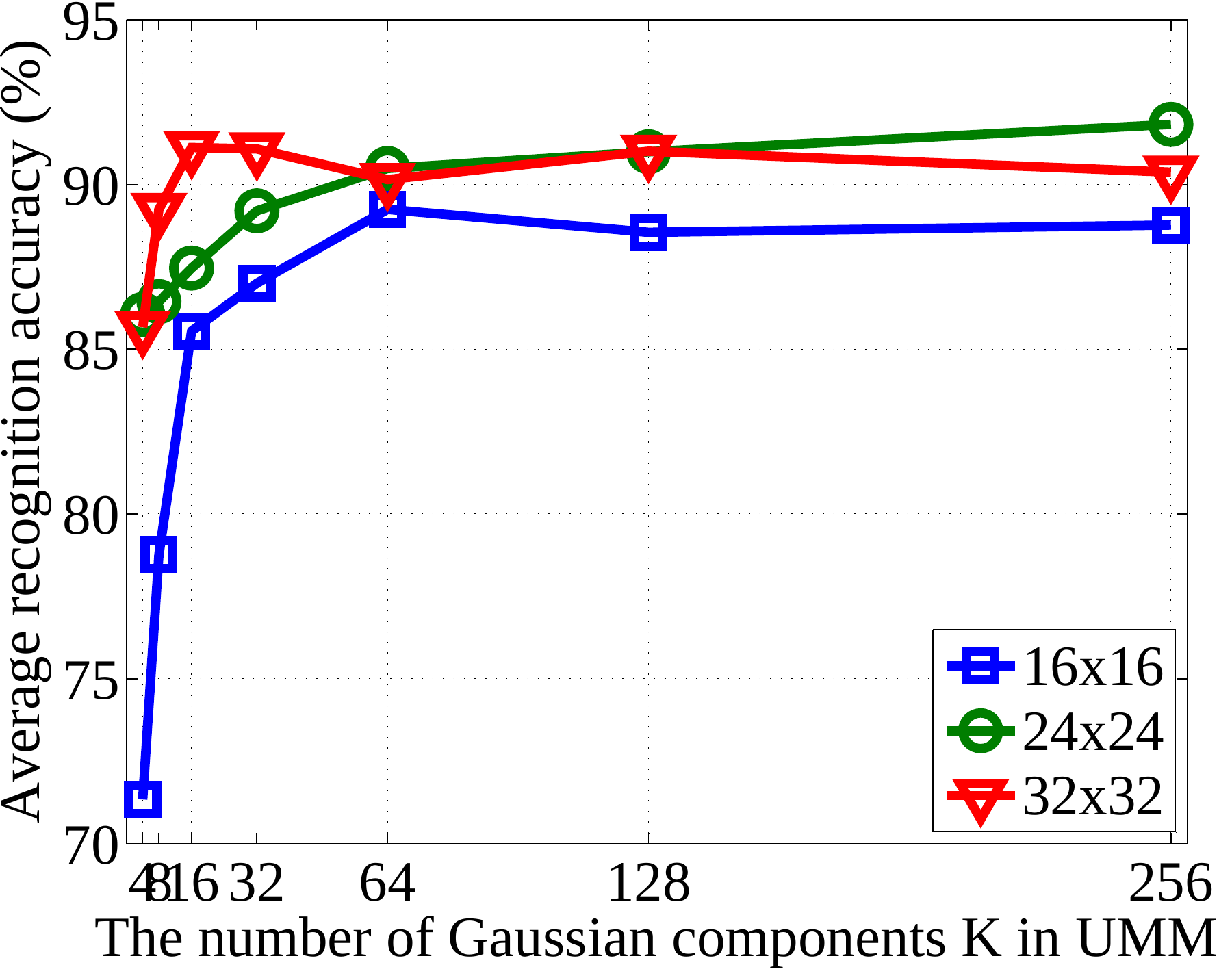}}
\subfigure[Oulu-CASIA]{
\includegraphics[height=3.2cm]{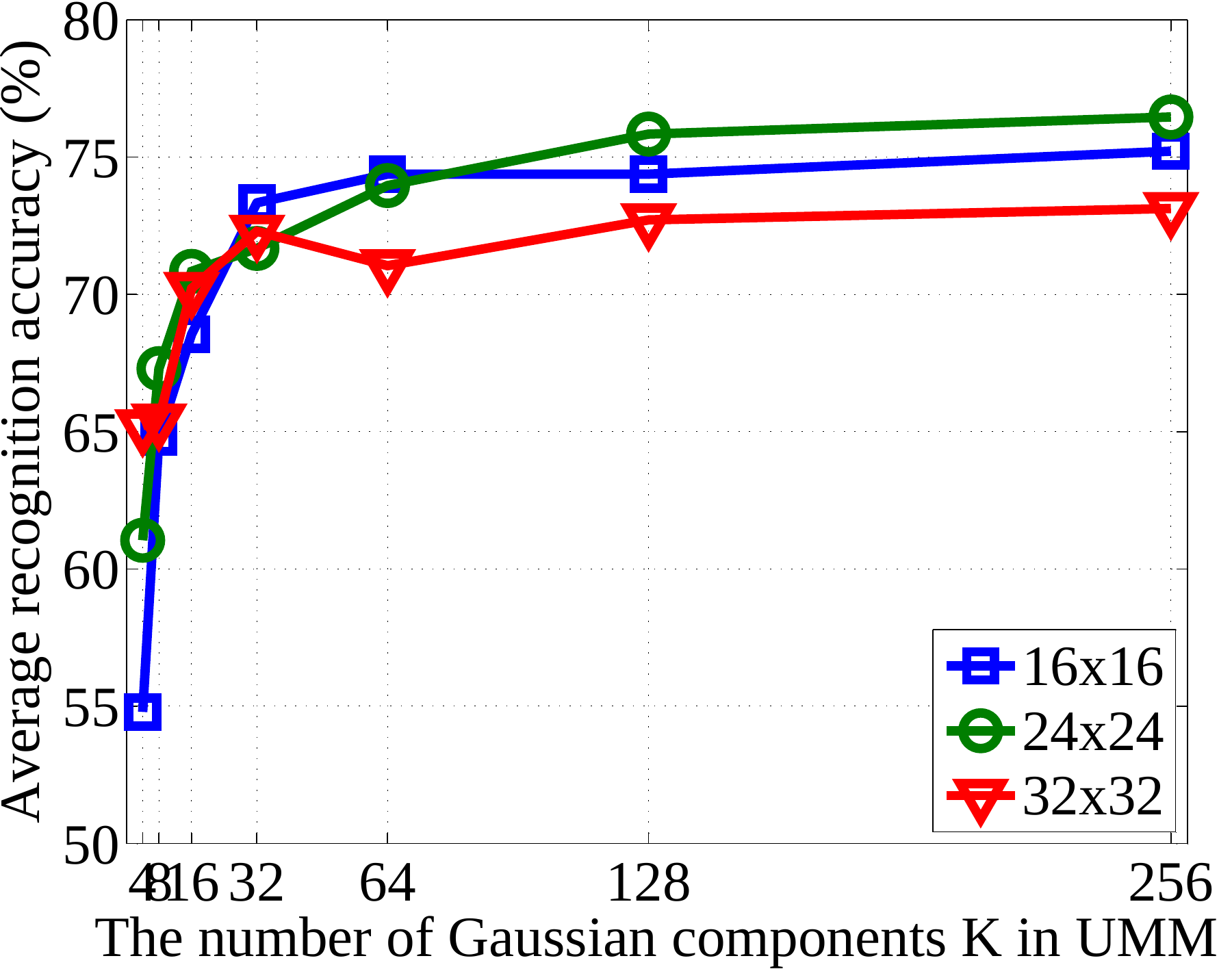}}
\subfigure[MMI]{
\includegraphics[height=3.2cm]{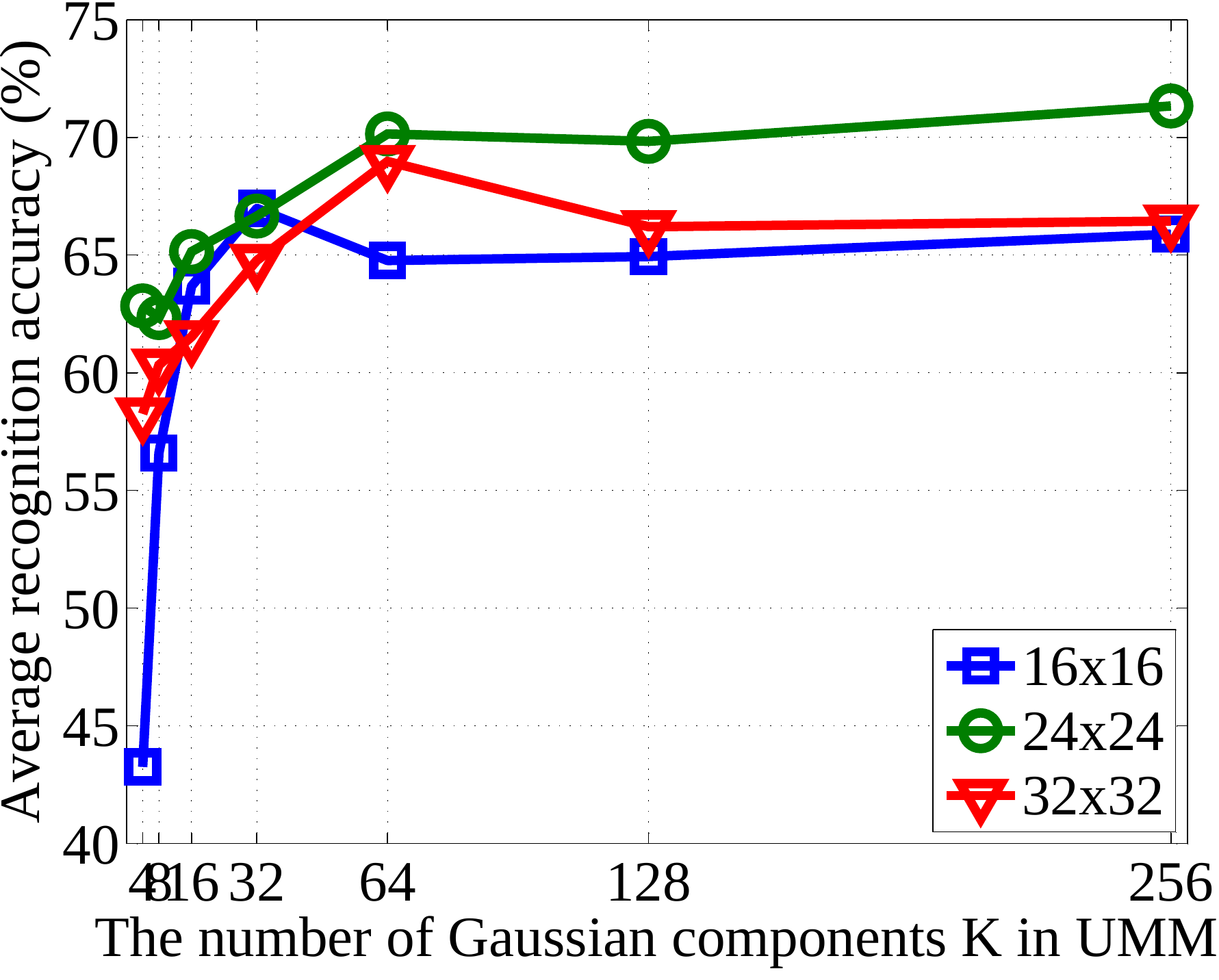}}
\subfigure[FERA]{
\includegraphics[height=3.2cm]{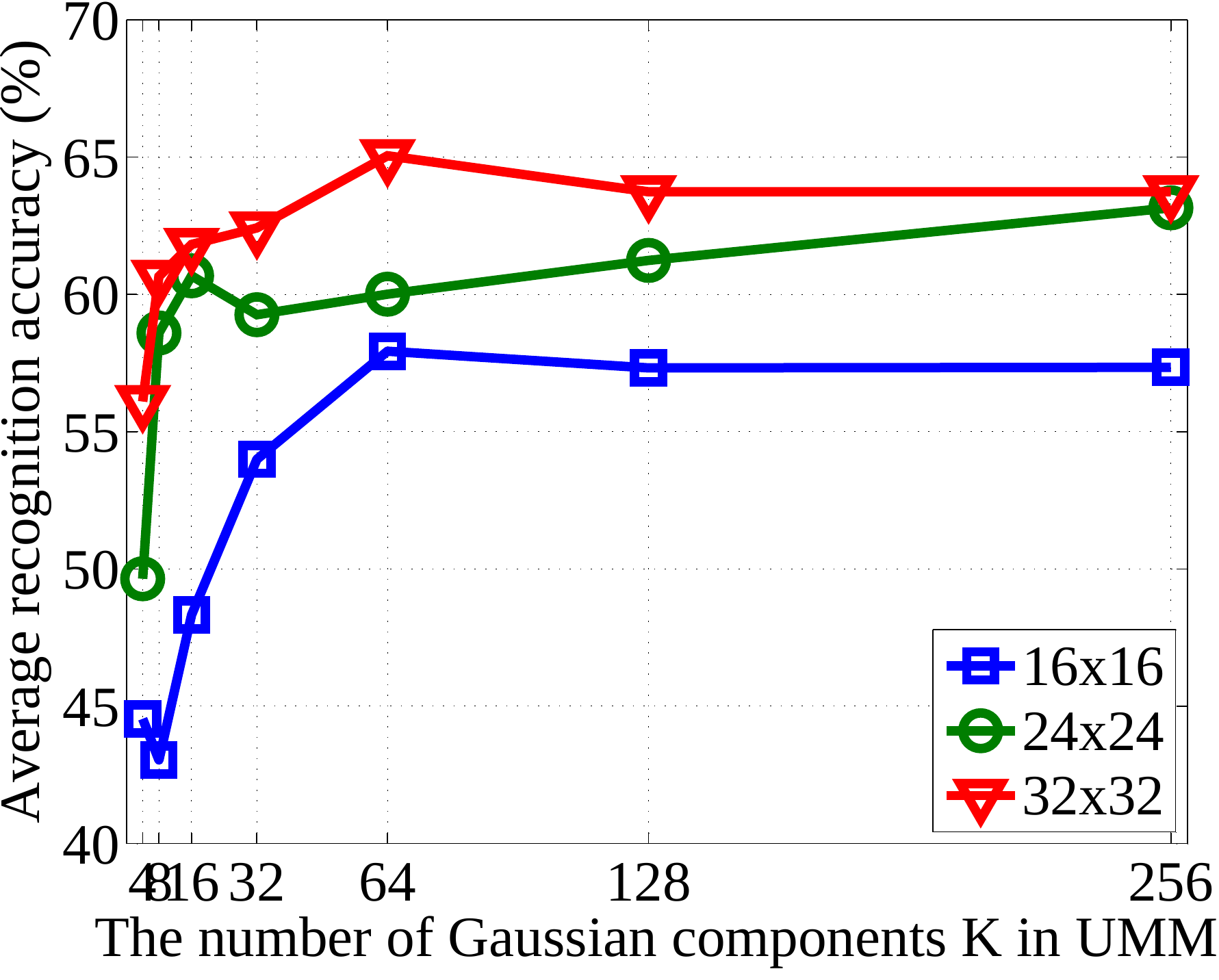}}
\caption{Average recognition accuracy (\%) with different patch sizes for low-level feature extraction on four datasets. (a) CK+ (b) Oulu-CASIA (c) MMI (d) FERA. (\textbf{using Dense SIFT feature}).}
\label{fig:figPatchsize}
\end{figure}

We first evaluate the effect of spatial scale, i.e. patch size $w,h$, for low-level feature extraction. The $w,h$ are varying in $16,24,32$. Here we only take SIFT feature for example. Other parameters $T=64$ and $dim=256$ are fixed in the experiments on all datasets. Figure~\ref{fig:figPatchsize} illustrates the performance of different patch sizes with different numbers of Gaussian components $K$. As shown, on CK+, Oulu-CASIA, MMI, the green curves with $24*24$ perform the best. While on FERA, the results become better when adopting larger patch size. The reason may be that muscle motions induced by spontaneous expression is likely to involve larger facial regions compared to posed expression. In the following evaluations, we uniformly apply $w=h=24$ on all datasets.

\subsubsection{The effect of alignment via UMM}

\begin{figure}[tbh]
\centering
\includegraphics[height=4.2cm]{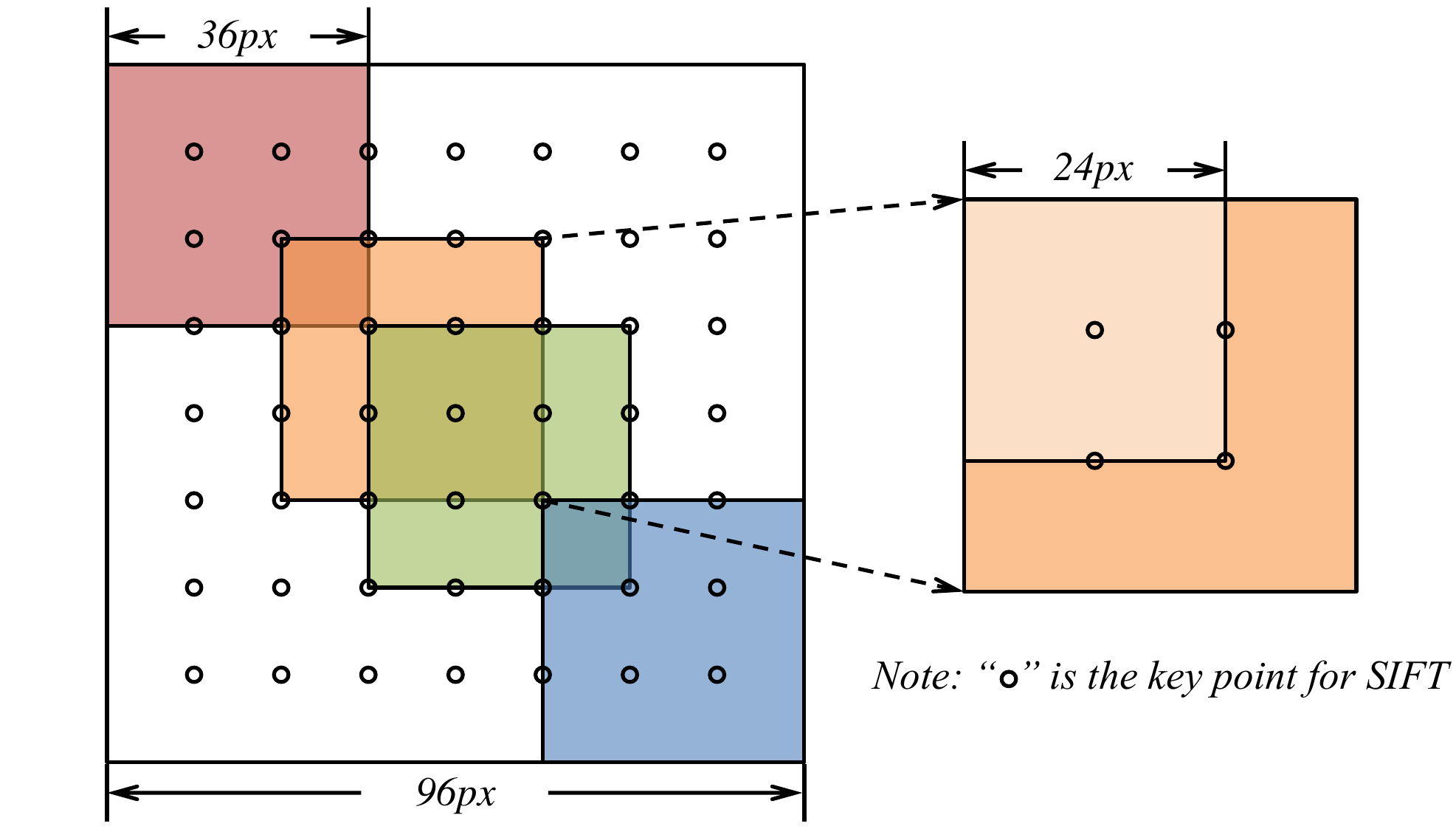}
\caption{An illustration of rigid blocking scheme in spatial dimension. The whole image is $96*96$ pixels and each block is $36*36$ pixels in spatial. For $w=h=24$, the whole image contains $7*7=49$ key points ``$\circ$'' for SIFT descriptor and each block covers 4 as shown in the right.}
\label{fig:figExpRigid}
\end{figure}

We compare the rigid blocking and elastic alignment (UMM) manners for the construction of a bank of local modes. In our experiments, the number of blocks/modes $K$ is varying in 16,32,64,128,256. For rigid blocking manner, the number of blocks in spatial dimension is fixed to $4*4=16$ and the blocking scheme is illustrated in Figure~\ref{fig:figExpRigid}. Then the number of partitions in temporal dimension is $K/16$ (i.e. 1,2,4,8,16).

\begin{figure}[tbh]
\centering
\subfigure[CK+]{
\includegraphics[height=3.2cm]{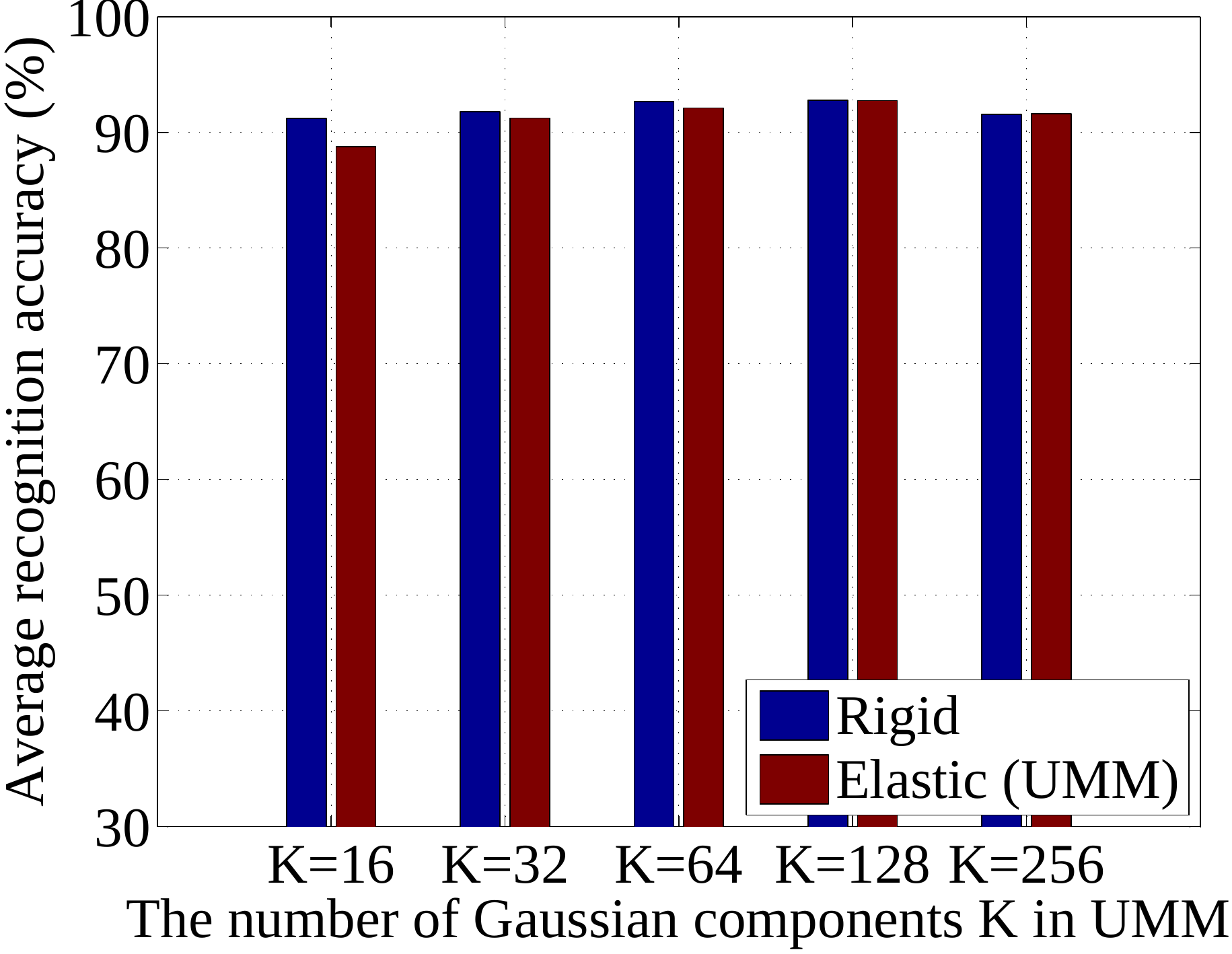}}
\subfigure[Oulu-CASIA]{
\includegraphics[height=3.2cm]{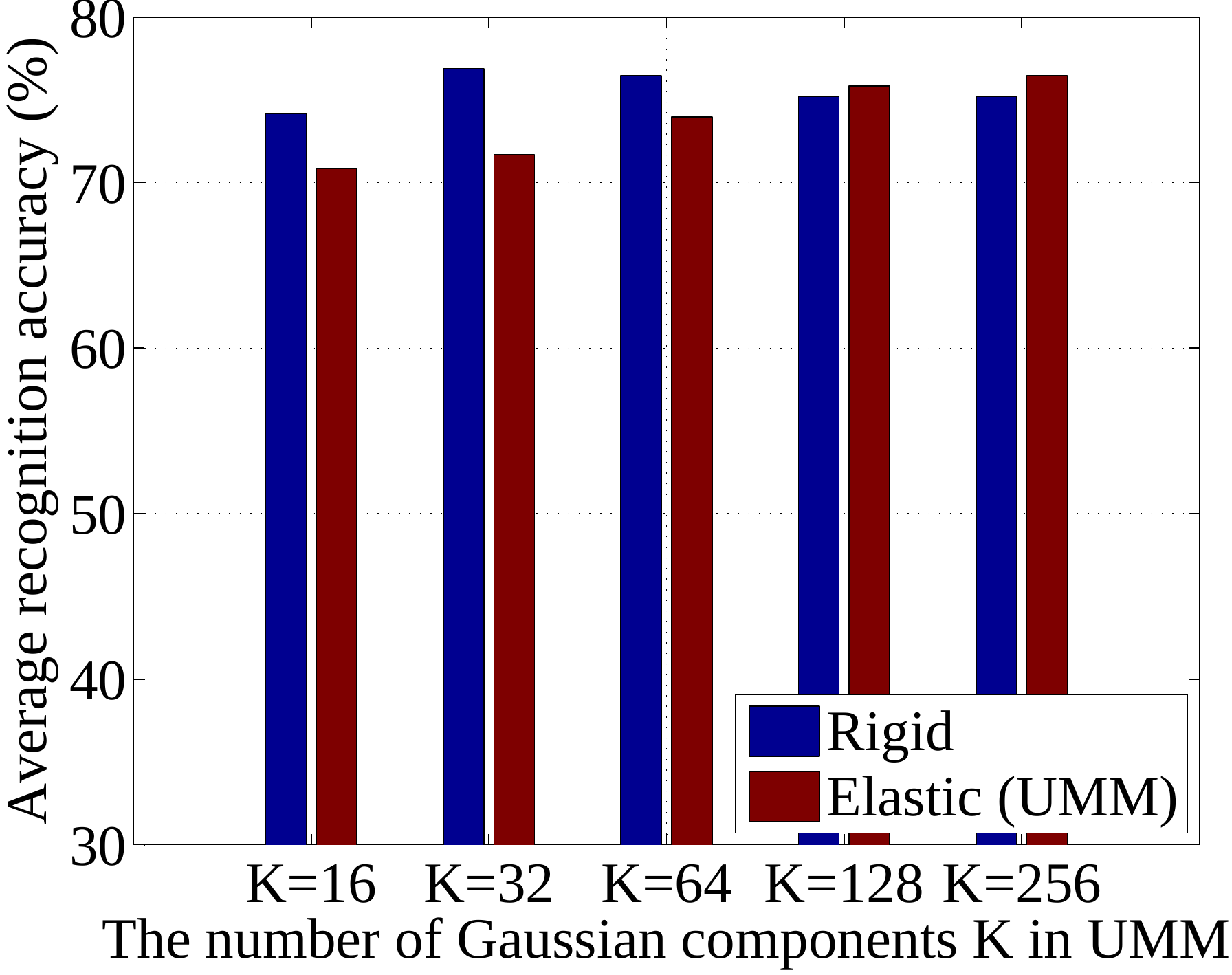}}
\subfigure[MMI]{
\includegraphics[height=3.2cm]{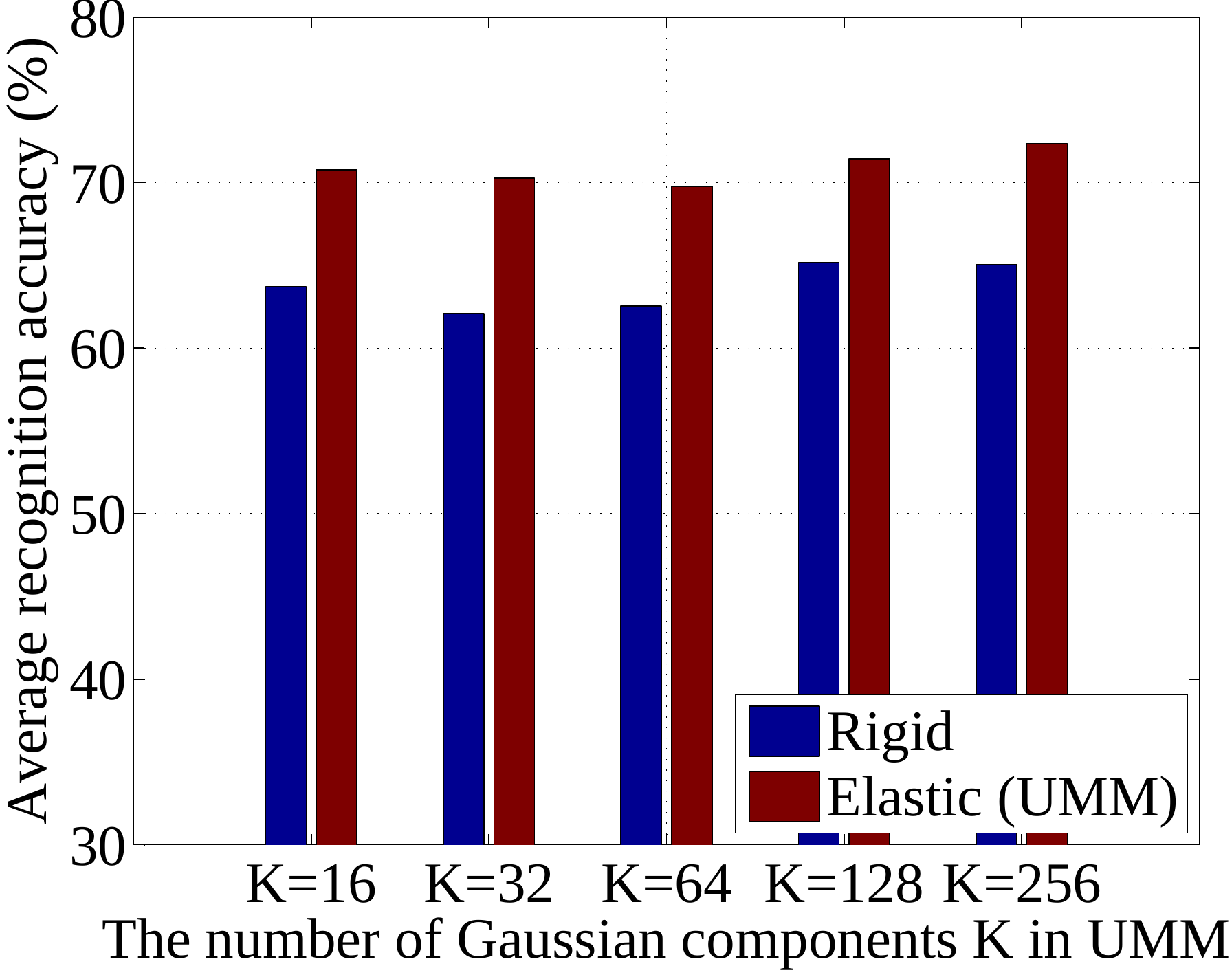}}
\subfigure[FERA]{
\includegraphics[height=3.2cm]{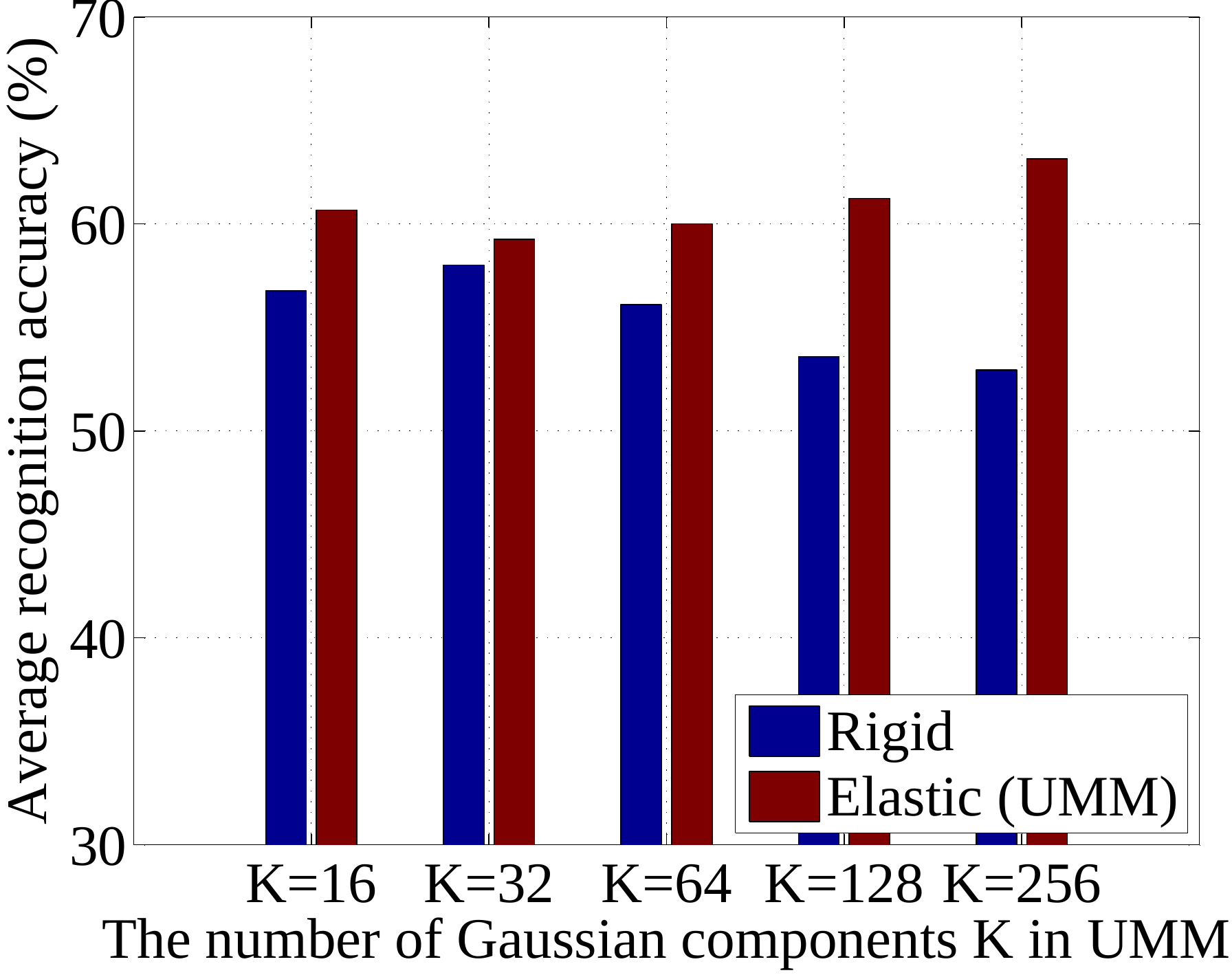}}
\caption{Average recognition accuracy (\%) with different alignment manners (rigid/elastic) on four datasets. (a) CK+ (b) Oulu-CASIA (c) MMI (d) FERA. (\textbf{using Dense SIFT feature}).}
\label{fig:figAlignment}
\end{figure}

The performance comparison is shown in Figure~\ref{fig:figAlignment}. On CK+ and Oulu-CASIA, the elastic manner performs not better than rigid manner, especially with smaller value of $K$ on Oulu-CASIA. It is possibly due to that the expression sequences of CK+ and Oulu-CASIA demonstrate a monotonous variation from neutral to apex status, thus the temporal alignment is not the major challenge for recognition. For MMI, each of the sequence reflects the whole temporal activation from onset to apex and then to offset of a single expression in a long term; For FERA, the expression samples show much more complex temporal variations in the spontaneous manner, even with no explicit segmentation of onset, apex, or offset stages. In such situation, a temporal alignment becomes crucial for building correspondence among different sequences. As verified in our experiments, the elastic manner performs much better than the rigid manner on MMI and FERA databases. It can be observed that the improvement becomes more significant as $K$ increases, which indicates that a larger number of local modes leads to a more elaborate alignment.

\subsubsection{The effect of low-level feature assignment manner}

\begin{figure}
\centering
\subfigure[CK+]{
\includegraphics[height=3.2cm]{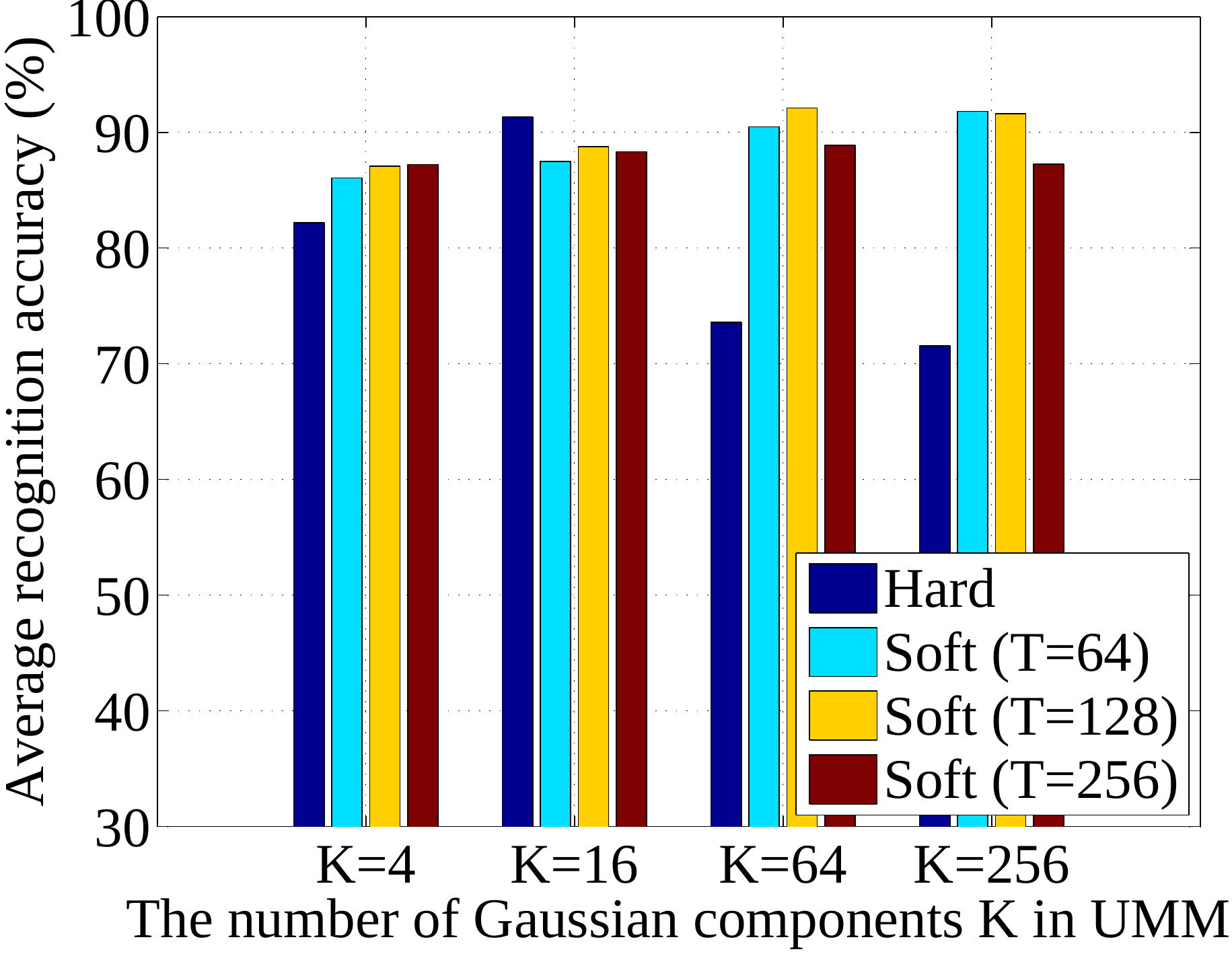}}
\subfigure[Oulu-CASIA]{
\includegraphics[height=3.2cm]{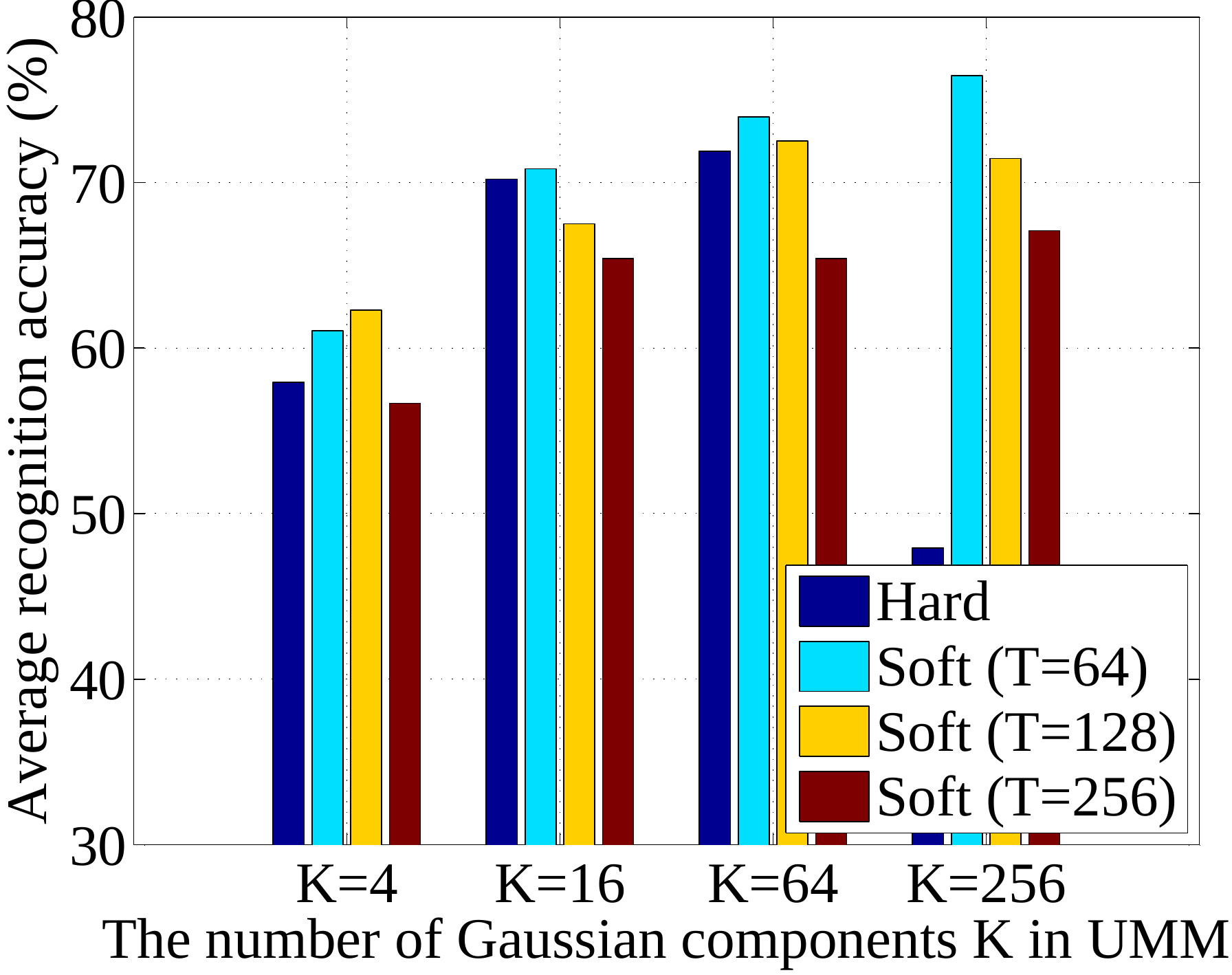}}
\subfigure[MMI]{
\includegraphics[height=3.2cm]{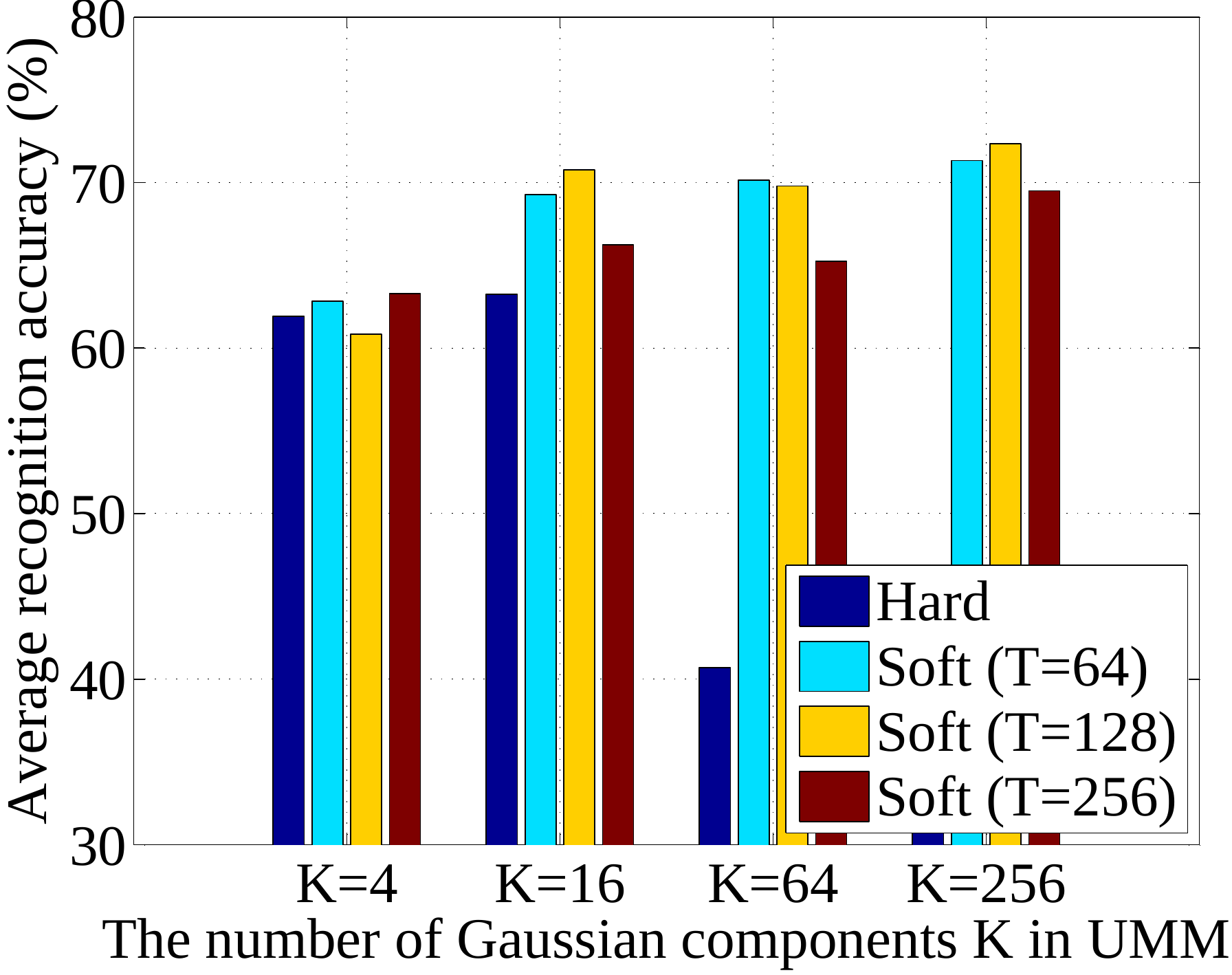}}
\subfigure[FERA]{
\includegraphics[height=3.2cm]{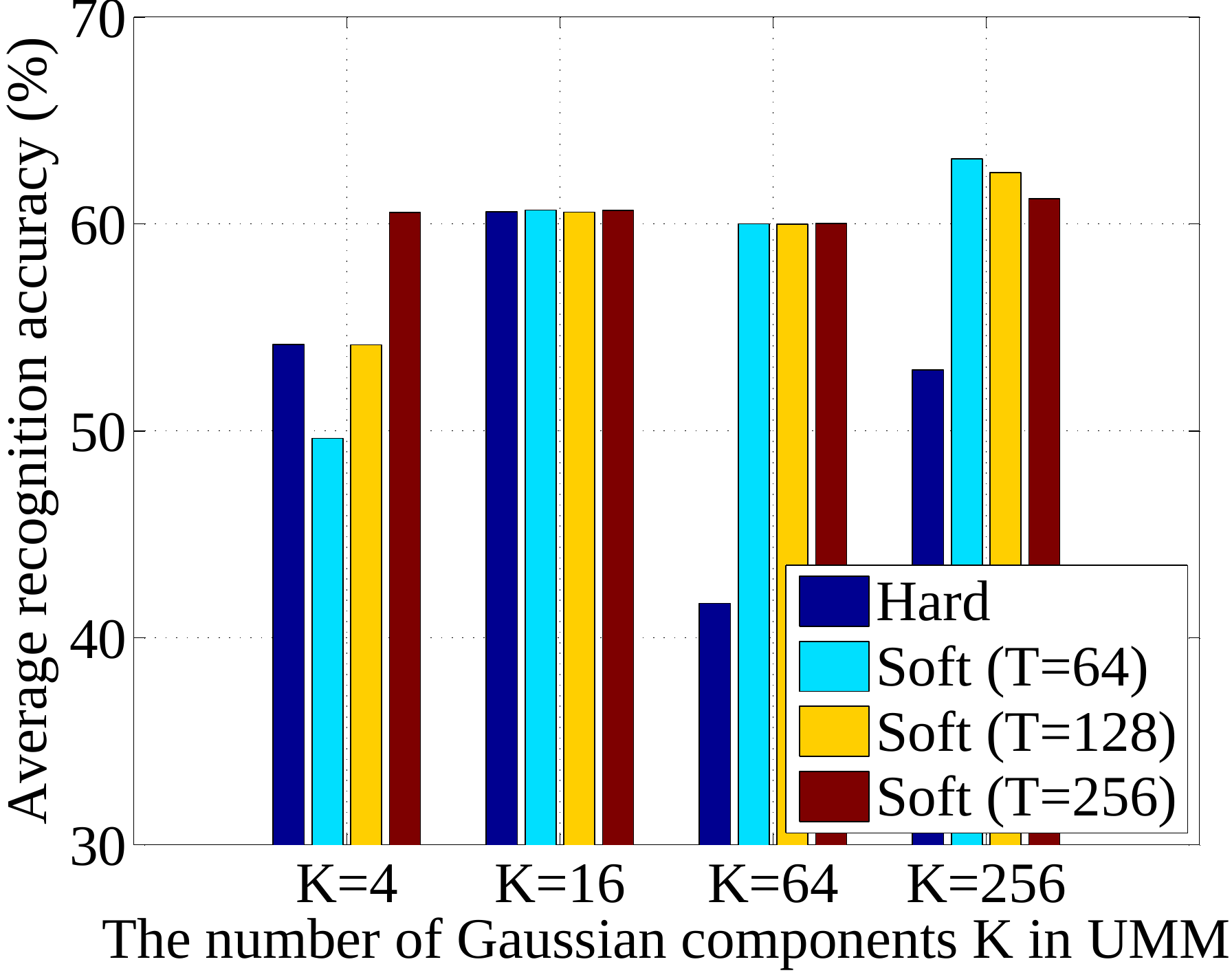}}
\caption{Average recognition accuracy (\%) with different assignment manners (hard/soft) on four datasets. (a) CK+ (b) Oulu-CASIA (c) MMI (d) FERA.}
\label{fig:figAssignment}
\vspace{-10pt}
\end{figure}

In UMM fitting stage, there are also two options for low-level feature assignment to each local mode (i.e. Gaussian component). For hard assignment, each low-level feature must be assigned to only one certain component according to its largest probability (i.e. traditional GMM). For soft assignment applied in our method, each component can obtain a fixed number of features with top $T$ probabilities. We compare these two different manners under different number of local modes (Gaussian components) $K=4,8,16,32,64,128,256$ and further discuss the effect of different values of $T=64,128,256$ in soft assignment. A comprehensive evaluation results are listed in Table~\ref{tab:tabAssignment}, with a graphical illustration in Figure~\ref{fig:figAssignment}.
\begin{table}[tbh]
\caption{Average recognition accuracy (\%) with different assignment manners (hard/soft) on four datasets. (a) CK+ (b) Oulu-CASIA (c) MMI (d) FERA. (\textbf{using Dense SIFT feature}).}
\centering
\subtable[CK+]{
\begin{tabular}{c|ccccccc}
  \hline\hline
  & k=4 & k=8 & k=16 & k=32 & k=64 & k=128 & k=256 \\
  \hline
  Hard &82.17&89.60&91.34&\underline{92.13}&73.57&62.30&71.55 \\
  Soft64~ &86.05&86.45&87.46&89.20&90.48&90.99&91.82 \\
  Soft128 &87.10&88.56&88.78&91.23&92.09&\textbf{\underline{92.75}}&91.61 \\
  Soft256 &87.21&87.14&88.33&86.47&88.86&87.79&87.25 \\
  \hline\hline
\end{tabular}
}
\subtable[Oulu-CASIA]{
\begin{tabular}{c|ccccccc}
  \hline\hline
  & k=4 & k=8 & k=16 & k=32 & k=64 & k=128 & k=256 \\
  \hline
  Hard &57.92&67.08&70.21&\textbf{\underline{77.29}}&71.88&47.50&47.92 \\
  Soft64~ &61.04&67.29&70.83&71.67&73.96&75.83&\underline{76.46} \\
  Soft128 &62.29&63.75&67.50&69.79&72.50&71.25&71.46 \\
  Soft256 &56.67&61.46&65.42&65.42&65.42&67.29&67.08 \\
  \hline\hline
\end{tabular}
}
\subtable[MMI]{
\begin{tabular}{c|ccccccc}
  \hline\hline
  & k=4 & k=8 & k=16 & k=32 & k=64 & k=128 & k=256 \\
  \hline
  Hard &61.93&63.94&63.26&\underline{64.76}&40.69&37.59&43.90 \\
  Soft64~ &62.84&62.34&69.28&66.52&70.14&69.83&71.33 \\
  Soft128 &60.84&71.07&70.77&70.27&69.78&71.42&\textbf{\underline{72.36}} \\
  Soft256 &63.28&67.23&66.23&64.41&65.23&67.67&69.50 \\
  \hline\hline
\end{tabular}
}
\subtable[FERA]{
\begin{tabular}{c|ccccccc}
  \hline\hline
  & k=4 & k=8 & k=16 & k=32 & k=64 & k=128 & k=256 \\
  \hline
  Hard &54.16&60.59&\underline{60.59}&58.09&41.66&38.04&52.95 \\
  Soft64~ &49.64&58.59&60.67&59.26&60.00&61.23&\textbf{\underline{63.15}} \\
  Soft128 &54.16&62.69&60.57&61.21&59.98&58.67&62.48 \\
  Soft256 &60.55&60.61&60.65&61.28&60.02&60.63&61.21 \\
  \hline\hline
\end{tabular}
}
\label{tab:tabAssignment}
\end{table}

As shown, the results based on hard manner can reach its peak at $K=16$ or $32$, and then suffer significant degradation as $K$ increases. It is because that in hard manner, the larger $K$ leads to the less features assigned to each component, which results in inaccurate estimation of the feature covariance for expressionlet representation. However, with a fixed number of features in each mode, the soft manner can hold the increasing trend as $K$ becomes larger. On the other hand, to consider the effect of different values of $T$, the larger $T$, i.e. the more features selected in each local mode, does not always yield better performance. The reason may be that more ``noise'' features with low probabilities are involved when applying a larger $T$.
\begin{table*}
\caption{Average recognition accuracy (\%) comparison with ExpLet or Dis-ExpLet on four datasets. (a) CK+ (b) Oulu-CASIA (c) MMI (d) FERA. (\textbf{using Dense SIFT feature}).}
\centering
\subtable[CK+]{
\begin{tabular}{c|ccc|ccc}
  \hline\hline
  \multirow{2}{*}{$dim$} & \multicolumn{3}{c|}{ExpLet} & \multicolumn{3}{c}{Dis-ExpLet} \\
  \cline{2-7}
  & k=64 & k=128 & k=256 & k=64 & k=128 & k=256 \\
  \hline
  64 &86.19&87.16&88.57&91.01&91.10&88.03 \\
  128 &89.28&89.76&89.93&92.84&\textbf{\underline{93.81}}&90.56 \\
  256 &90.48&90.99&\underline{91.82}&92.81&93.34&93.05 \\
  \hline\hline
\end{tabular}
}
\hspace{10pt}
\subtable[Oulu-CASIA]{
\begin{tabular}{c|ccc|ccc}
  \hline\hline
  \multirow{2}{*}{$dim$} & \multicolumn{3}{c|}{ExpLet} & \multicolumn{3}{c}{Dis-ExpLet} \\
  \cline{2-7}
  & k=64 & k=128 & k=256 & k=64 & k=128 & k=256 \\
  \hline
  64 &71.04&70.00&72.08&73.13&76.46&74.79 \\
  128 &72.71&72.50&74.79&75.63&75.83&77.50 \\
  256 &73.96&75.83&\underline{76.46}&76.46&77.71&\textbf{\underline{78.96}} \\
  \hline\hline
\end{tabular}
}
\subtable[MMI]{
\begin{tabular}{c|ccc|ccc}
  \hline\hline
  \multirow{2}{*}{$dim$} & \multicolumn{3}{c|}{ExpLet} & \multicolumn{3}{c}{Dis-ExpLet} \\
  \cline{2-7}
  & k=64 & k=128 & k=256 & k=64 & k=128 & k=256 \\
  \hline
  64 &61.55&64.53&68.06&76.56&72.61&74.30 \\
  128 &68.56&67.51&68.15&\textbf{\underline{76.65}}&73.79&74.93 \\
  256 &70.14&69.83&\underline{71.33}&76.60&75.57&76.51 \\
  \hline\hline
\end{tabular}
}
\hspace{10pt}
\subtable[FERA]{
\begin{tabular}{c|ccc|ccc}
  \hline\hline
  \multirow{2}{*}{~$dim$} & \multicolumn{3}{c|}{ExpLet} & \multicolumn{3}{c}{Dis-ExpLet} \\
  \cline{2-7}
  & k=64 & k=128 & k=256 & k=64 & k=128 & k=256 \\
  \hline
  64 &60.03&58.07&61.25&54.18&64.38&65.00 \\
  128 &59.98&60.61&61.23&63.29&70.27&70.18 \\
  256 &60.00&61.23&\underline{63.15}&64.48&\textbf{\underline{72.91}}&68.41 \\
  \hline\hline
\end{tabular}
}
\label{tab:tabDisExpLet}
\end{table*}
\begin{figure*}
\centering
\subfigure[CK+]{
\includegraphics[height=3.8cm]{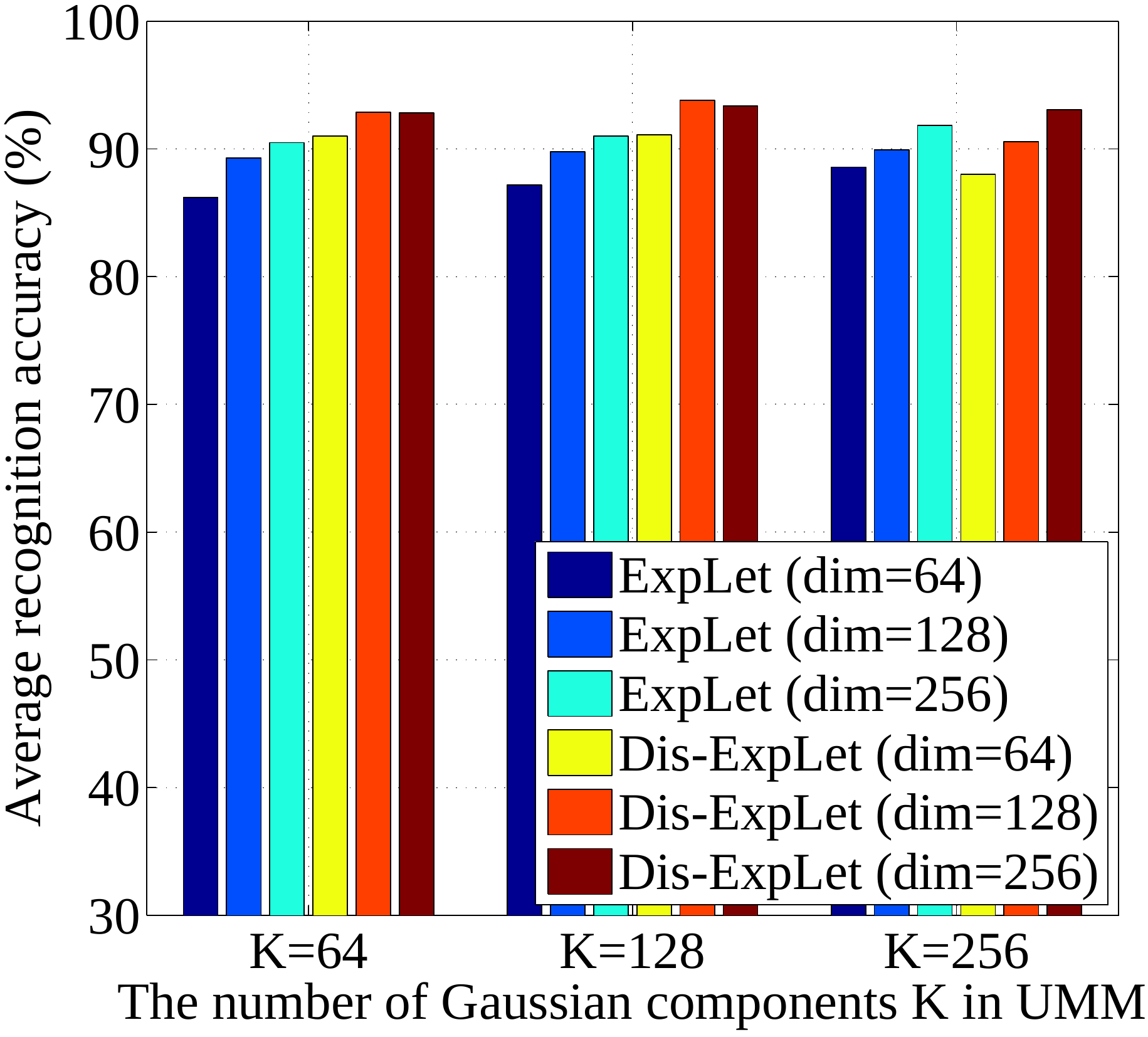}}
\subfigure[Oulu-CASIA]{
\includegraphics[height=3.8cm]{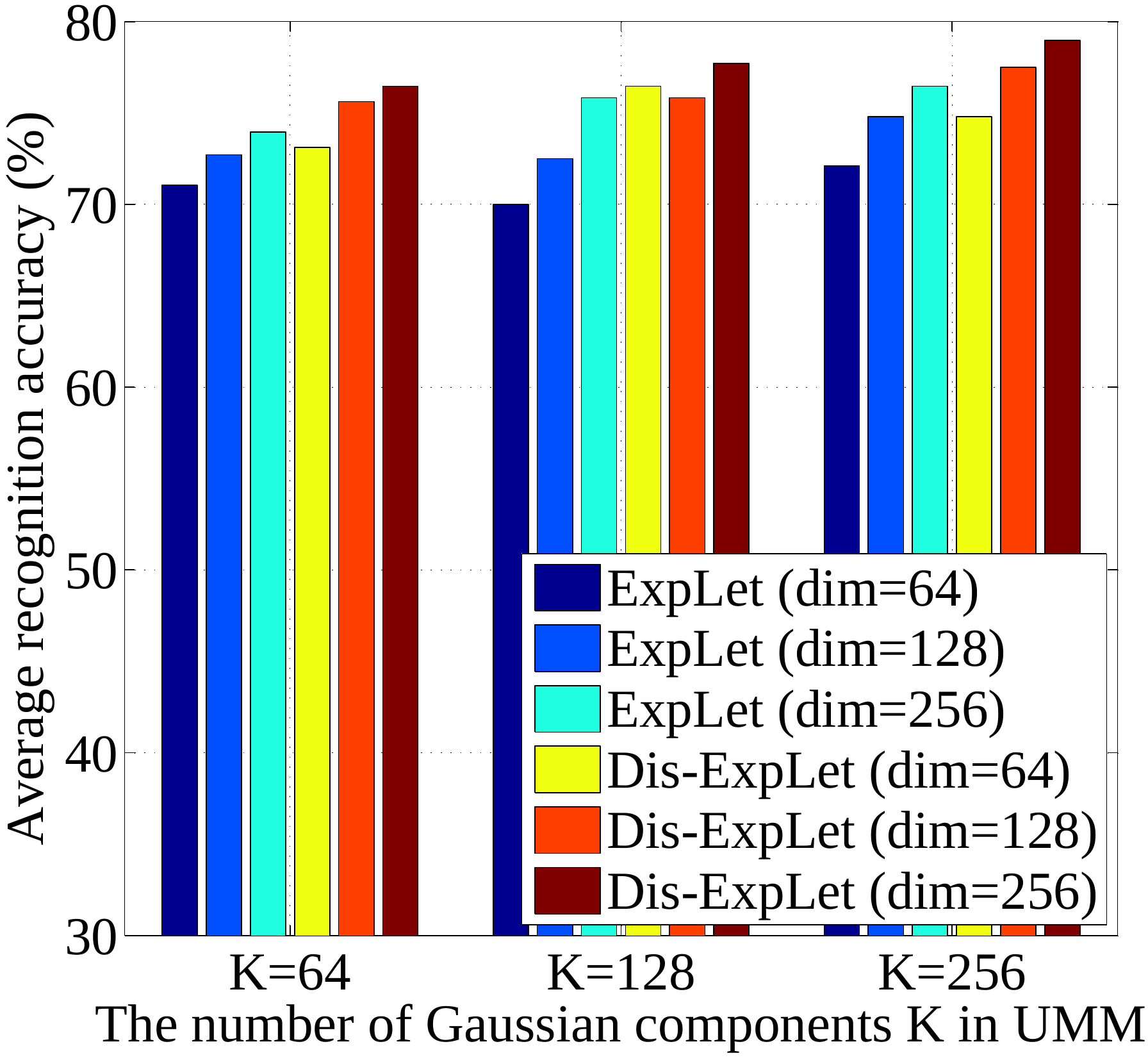}}
\subfigure[MMI]{
\includegraphics[height=3.8cm]{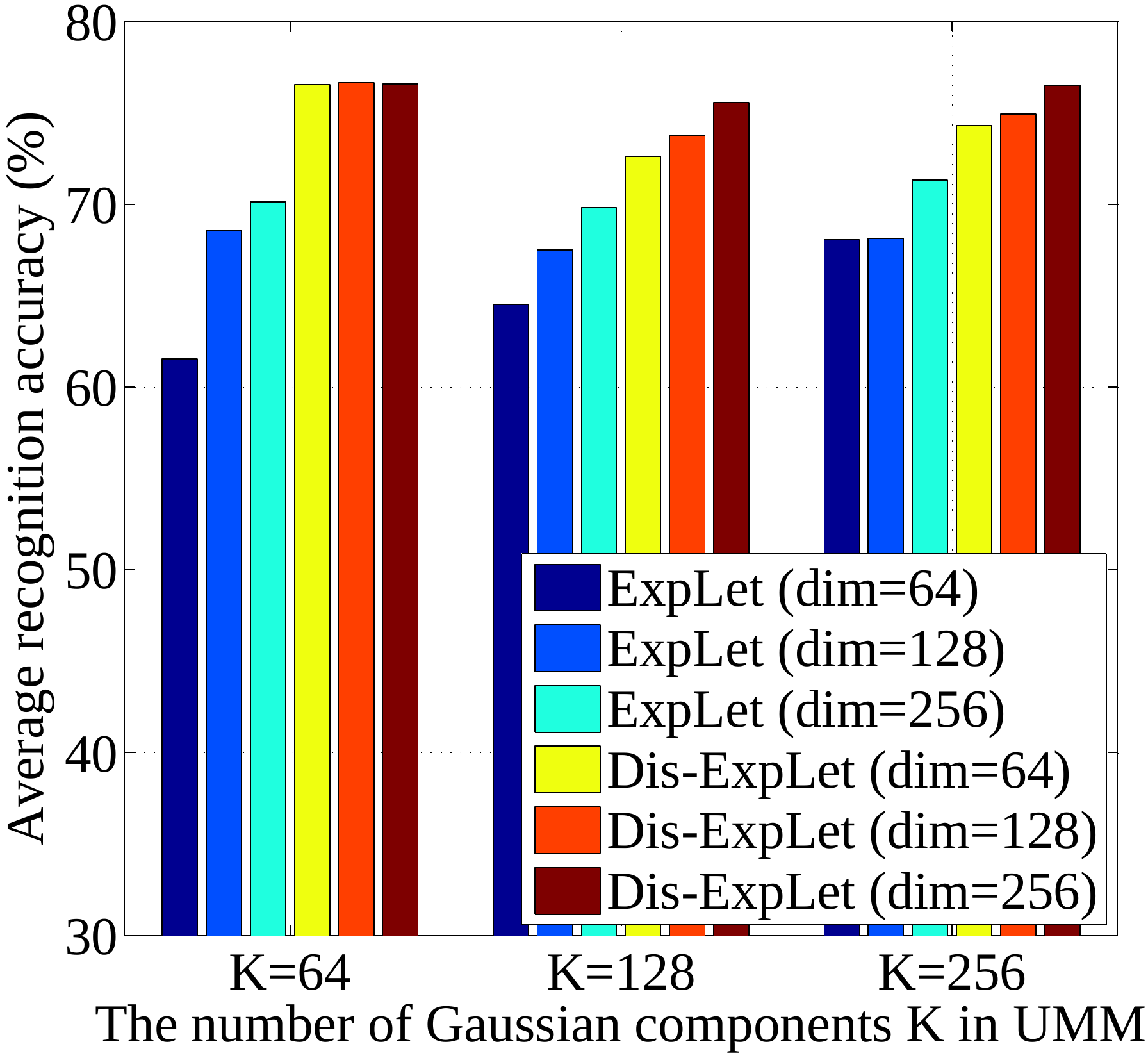}}
\subfigure[FERA]{
\includegraphics[height=3.8cm]{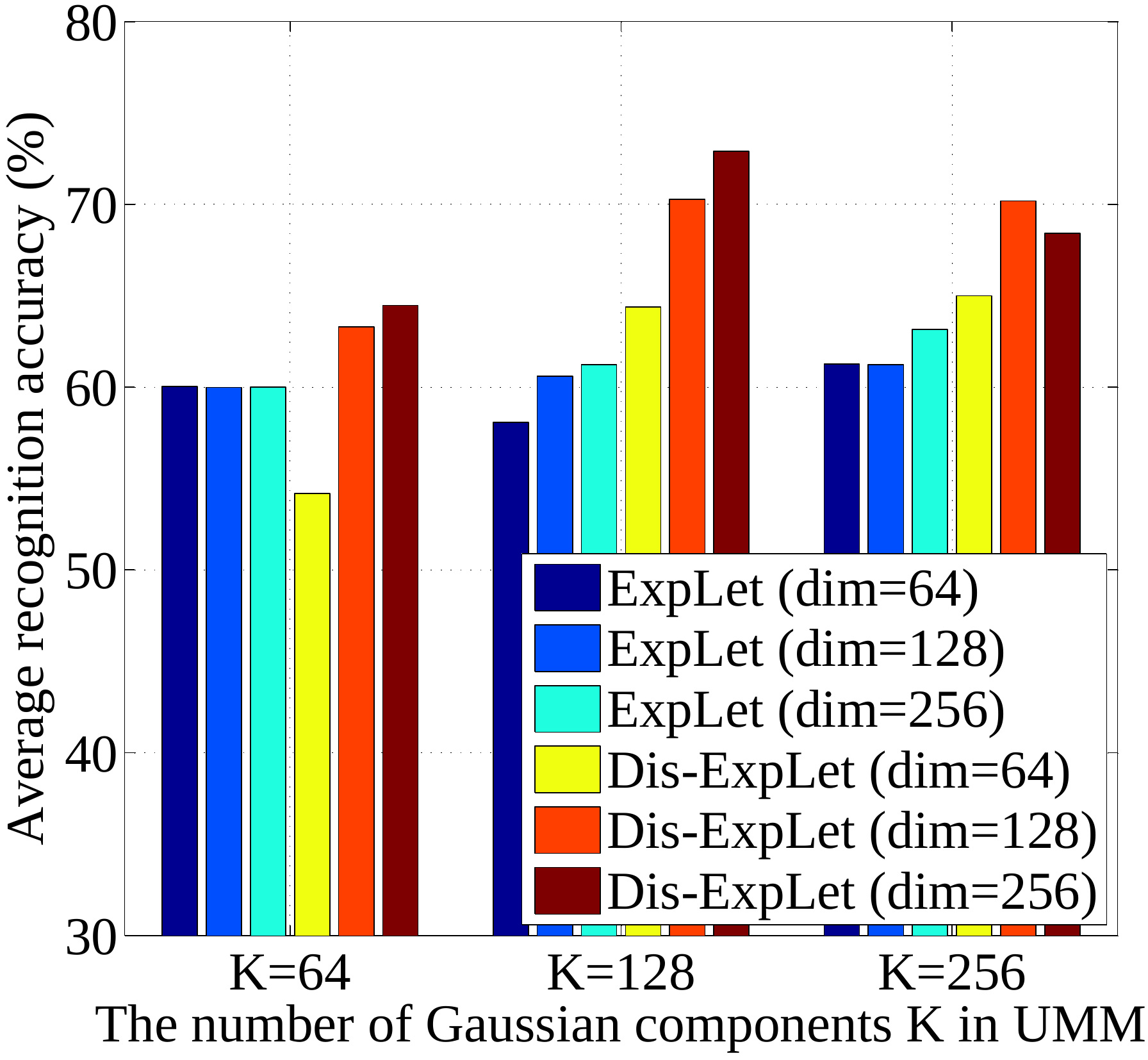}}
\caption{Average recognition accuracy (\%) comparison with ExpLet or Dis-ExpLet on four datasets. (a) CK+ (b) Oulu-CASIA (c) MMI (d) FERA.}
\label{fig:figDisExpLet}
\end{figure*}

\vspace{5pt}
\subsubsection{The effect of discriminant learning}

Finally we evaluate the effect of discriminant learning with expressionlets. The original dimension of expressionlets is $64*64=4096d$ as the low-level features are $64d$. For dimension reduction, we can simply apply unsupervised PCA or employ the proposed marginal discriminant learning. Here we compare these two schemes by varying $dim=64,128,256$ under different $K$, and the results are shown in Table~\ref{tab:tabDisExpLet} and Figure~\ref{fig:figDisExpLet}. It can be observed that ``Dis-ExpLet'' performs much better than ``ExpLet'' even using a lower dimension. The improvement is quite significant especially on MMI ($\sim5.3\%$) and FERA ($\sim9.7\%$), which are considered to be more challenging than CK+ and Oulu-CASIA.

\subsection{Comprehensive comparisons with Fisher Vector}

In this section, we conduct comprehensive comparisons with the state-of-the-art encoding method Fisher Vector. The experiments are conducted based on two kinds of descriptors, i.e. SIFT (2D) and HOG (3D). And for Fisher Vector, we also tune different values of $w,h$ to obtain the best performance. All of the results are listed in Table~\ref{tab:tabFV}.

According to the results, for $w=h=16$ or $24$, we can always observe a approximately rising trend of accuracy as the number of GMM components $K$ increases. However, for $w=h=32$, there usually exist an obvious degradation as $K$ increases (except for Oulu-CASIA). It may be caused by that the patches with a larger scale encode less details which cannot provide enough local patterns for lots of partitions. Thus when $K$ becomes larger, the cluster partitions forcibly segment some similar or related patterns, which brings confusions in pooling stage for higher-level semantics generation.

For fair comparison, in Table~\ref{tab:tabExpLet} we report the performance based on original ``ExpLet'' (the dimension is reduced to $dim$ via unsupervised PCA) without discriminant learning. Here we fix the other parameters $w=h=24$ and $T=64$. As shown, the performance improves gradually with the increasing of the number of ``ExpLet'' $K$ and the preserved dimension $dim$, and the peak values outperform the FV results significantly. Even with the same dimension of final FV representation (i.e. $2*64*k=128k$), our method (with $dim=128$) always performs a little better, which proves that the covariance pooling scheme can capture more dynamic information for expression description thus benefits the final recognition.

Another observation is about the results based on different descriptors. For both FV and ExpLet, on CK+, Oulu-CASIA, and MMI, dense SIFT consistently performs much better than HOG, while on FERA, the HOG shows clearly superior to SIFT under all settings. The main difference of the two descriptors is whether encoding the temporal information, i.e. SIFT is in 2D and HOG is in 3D. We conjecture that for spontaneous samples in FERA, the variations along temporal dimension are more complex and thus require more detailed and elaborate encoding via low-level descriptors.
\begin{table*}
\caption{Average recognition accuracy (\%) based on Fisher Vector on four databases.}
\centering
\subtable[CK+ (HOG)]{
\begin{tabular}{c|ccccccc}
  \hline\hline
  $w,h$ & k=4 & k=8 & k=16 & k=32 & k=64 & k=128 & k=256 \\
  \hline
  16 &58.07&65.91&68.43&78.27&77.10&80.22&83.60 \\
  24 &61.17&71.10&79.05&82.42&82.92&\textbf{\underline{84.31}}&82.88 \\
  32 &66.36&73.16&76.68&81.52&80.83&80.56&78.38 \\
  \hline\hline
\end{tabular}
}
\subtable[CK+ (SIFT)]{
\begin{tabular}{c|ccccccc}
  \hline\hline
  $w,h$ & k=4 & k=8 & k=16 & k=32 & k=64 & k=128 & k=256 \\
  \hline
  16 &69.35&75.24&81.14&85.04&85.09&83.55&84.69 \\
  24 &72.45&78.57&84.69&86.24&87.51&87.35&\textbf{\underline{89.18}} \\
  32 &77.02&84.92&83.64&84.99&88.88&85.95&83.44 \\
  \hline\hline
\end{tabular}
}
\centering
\subtable[Oulu-CASIA (HOG)]{
\begin{tabular}{c|ccccccc}
  \hline\hline
  $w,h$ & k=4 & k=8 & k=16 & k=32 & k=64 & k=128 & k=256 \\
  \hline
  16 &48.54&50.83&54.79&57.08&59.38&64.17&67.92 \\
  24 &51.88&57.08&62.71&61.67&67.08&67.71&\textbf{\underline{69.79}} \\
  32 &46.46&51.46&56.46&61.04&64.38&68.54&67.71 \\
  \hline\hline
\end{tabular}
}
\subtable[Oulu-CASIA (SIFT)]{
\begin{tabular}{c|ccccccc}
  \hline\hline
  $w,h$ & k=4 & k=8 & k=16 & k=32 & k=64 & k=128 & k=256 \\
  \hline
  16 &54.38&58.33&62.50&65.63&66.67&71.88&71.46 \\
  24 &49.58&59.79&61.67&62.08&68.13&68.96&71.67 \\
  32 &55.42&58.33&61.25&66.67&68.96&70.63&\textbf{\underline{72.92}} \\
  \hline\hline
\end{tabular}
}
\centering
\subtable[MMI (HOG)]{
\begin{tabular}{c|ccccccc}
  \hline\hline
  $w,h$ & k=4 & k=8 & k=16 & k=32 & k=64 & k=128 & k=256 \\
  \hline
  16 &42.06&47.02&48.33&55.57&55.10&54.00&58.58 \\
  24 &42.76&54.81&60.39&62.56&63.41&62.70&\textbf{\underline{64.54}} \\
  32 &37.99&50.94&57.37&59.24&63.99&64.03&53.77 \\
  \hline\hline
\end{tabular}
}
\subtable[MMI (SIFT)]{
\begin{tabular}{c|ccccccc}
  \hline\hline
  $w,h$ & k=4 & k=8 & k=16 & k=32 & k=64 & k=128 & k=256 \\
  \hline
  16 &40.06&58.42&54.00&60.49&63.39&66.32&65.11 \\
  24 &43.21&61.40&62.27&62.16&65.62&63.77&63.48 \\
  32 &49.33&57.32&62.33&66.28&\textbf{\underline{68.64}}&61.49&58.59 \\
  \hline\hline
\end{tabular}
}
\centering
\subtable[FERA (HOG)]{
\begin{tabular}{c|ccccccc}
  \hline\hline
  $w,h$ & k=4 & k=8 & k=16 & k=32 & k=64 & k=128 & k=256 \\
  \hline
  16 &53.00&58.88&53.06&55.02&61.45&60.83&58.19 \\
  24 &57.42&55.44&62.01&63.39&63.48&\textbf{\underline{67.29}}&66.58 \\
  32 &59.48&62.71&60.79&59.50&67.10&60.03&54.20 \\
  \hline\hline
\end{tabular}
}
\subtable[FERA (SIFT)]{
\begin{tabular}{c|ccccccc}
  \hline\hline
  $w,h$ & k=4 & k=8 & k=16 & k=32 & k=64 & k=128 & k=256 \\
  \hline
  16 &50.47&56.25&52.38&58.88&58.31&59.54&\textbf{\underline{62.15}} \\
  24 &59.69&56.06&61.33&60.17&60.83&60.17&60.67 \\
  32 &53.03&55.57&52.97&56.94&60.75&60.71&48.41 \\
  \hline\hline
\end{tabular}
}
\label{tab:tabFV}
\caption{Average recognition accuracy (\%) based on Expressionlet on four databases.}
\centering
\subtable[CK+ (HOG)]{
\begin{tabular}{c|ccccccc}
  \hline\hline
  $dim$ & k=4 & k=8 & k=16 & k=32 & k=64 & k=128 & k=256 \\
  \hline
  32  &60.05&68.25&75.35&72.18&77.81&77.92&77.14 \\
  64  &65.57&70.67&73.51&80.55&79.52&78.55&80.39 \\
  128 &72.66&77.03&77.04&80.26&82.27&81.68&81.19 \\
  256 &76.26&77.00&81.09&81.70&85.45&82.97&82.37 \\
  512 &76.38&77.17&83.19&83.46&\textbf{\underline{85.56}}&83.77&82.82 \\
  \hline\hline
\end{tabular}
}
\subtable[CK+ (SIFT)]{
\begin{tabular}{c|ccccccc}
  \hline\hline
  $dim$ & k=4 & k=8 & k=16 & k=32 & k=64 & k=128 & k=256 \\
  \hline
  32  &68.09&75.11&78.69&82.31&80.71&85.51&83.28 \\
  64  &76.59&78.59&84.48&86.72&86.19&87.16&88.57 \\
  128 &82.30&84.42&84.16&88.16&89.28&89.76&89.93 \\
  256 &86.05&86.45&87.46&89.20&90.48&90.99&\textbf{\underline{91.82}} \\
  512 &87.06&88.02&87.15&90.04&90.20&90.99&90.71 \\
  \hline\hline
\end{tabular}
}
\centering
\subtable[Oulu-CASIA (HOG)]{
\begin{tabular}{c|ccccccc}
  \hline\hline
  $dim$ & k=4 & k=8 & k=16 & k=32 & k=64 & k=128 & k=256 \\
  \hline
  32  &33.75&41.04&50.63&57.29&60.63&61.67&63.75 \\
  64  &36.88&48.96&60.21&62.08&63.96&64.79&67.08 \\
  128 &45.21&54.79&66.67&65.83&66.46&70.21&69.58 \\
  256 &50.63&58.54&68.13&68.75&70.21&72.08&72.50 \\
  512 &54.17&62.50&69.38&71.04&72.50&73.54&\textbf{\underline{73.75}} \\
  \hline\hline
\end{tabular}
}
\subtable[Oulu-CASIA (SIFT)]{
\begin{tabular}{c|ccccccc}
  \hline\hline
  $dim$ & k=4 & k=8 & k=16 & k=32 & k=64 & k=128 & k=256 \\
  \hline
  32  &42.08&53.54&58.33&60.42&66.88&66.46&69.38 \\
  64  &51.25&61.46&65.42&64.38&71.04&70.00&72.08 \\
  128 &57.50&64.79&66.25&69.17&72.71&72.50&74.79 \\
  256 &61.04&67.29&70.83&71.67&73.96&75.83&76.46 \\
  512 &63.96&69.38&73.75&73.75&75.63&76.46&\textbf{\underline{76.88}} \\
  \hline\hline
\end{tabular}
}
\centering
\subtable[MMI (HOG)]{
\begin{tabular}{c|ccccccc}
  \hline\hline
  $dim$ & k=4 & k=8 & k=16 & k=32 & k=64 & k=128 & k=256 \\
  \hline
  32  &37.45&42.75&51.88&58.70&64.04&58.02&61.25 \\
  64  &42.22&53.90&56.76&62.52&66.44&62.16&64.30 \\
  128 &46.47&55.58&60.41&65.05&68.64&63.64&\textbf{\underline{69.18}} \\
  256 &49.38&58.92&62.35&66.53&69.02&63.91&68.92 \\
  512 &49.11&58.96&63.16&67.00&67.58&66.33&68.32 \\
  \hline\hline
\end{tabular}
}
\subtable[MMI (SIFT)]{
\begin{tabular}{c|ccccccc}
  \hline\hline
  $dim$ & k=4 & k=8 & k=16 & k=32 & k=64 & k=128 & k=256 \\
  \hline
  32  &46.13&49.12&51.71&58.42&63.89&63.27&65.16 \\
  64  &52.49&55.16&58.86&67.07&61.55&64.53&68.06 \\
  128 &59.59&59.00&65.33&66.84&68.56&67.51&68.15 \\
  256 &62.84&62.34&65.15&66.66&70.14&69.83&71.33 \\
  512 &62.16&63.83&69.35&68.17&\textbf{\underline{72.00}}&71.29&71.88 \\
  \hline\hline
\end{tabular}
}
\centering
\subtable[FERA (HOG)]{
\begin{tabular}{c|ccccccc}
  \hline\hline
  $dim$ & k=4 & k=8 & k=16 & k=32 & k=64 & k=128 & k=256 \\
  \hline
  32  &51.99&43.06&52.30&61.32&59.90&63.75&61.85 \\
  64  &53.37&44.41&54.84&59.34&61.88&65.06&63.79 \\
  128 &57.95&49.60&58.69&61.92&63.79&67.58&68.25 \\
  256 &56.53&52.18&58.05&62.50&63.77&68.20&\textbf{\underline{69.52}} \\
  512 &56.53&50.91&58.01&62.46&61.83&65.62&67.58 \\
  \hline\hline
\end{tabular}
}
\subtable[FERA (SIFT)]{
\begin{tabular}{c|ccccccc}
  \hline\hline
  $dim$ & k=4 & k=8 & k=16 & k=32 & k=64 & k=128 & k=256 \\
  \hline
  32  &38.75&50.28&54.14&56.72&60.65&\textbf{\underline{64.50}}&62.52 \\
  64  &41.35&56.66&56.15&54.84&60.03&58.07&61.25 \\
  128 &45.79&58.55&56.80&55.49&59.98&60.61&61.23 \\
  256 &49.64&58.59&60.67&59.26&60.00&61.23&63.15 \\
  512 &48.99&59.22&63.84&60.57&63.19&62.50&63.77 \\
  \hline\hline
\end{tabular}
}
\label{tab:tabExpLet}
\end{table*}

\subsection{Comparisons with state-of-the-art methods}

\begin{figure*}
\centering
\subfigure[CK+]{
\includegraphics[height=2.5cm]{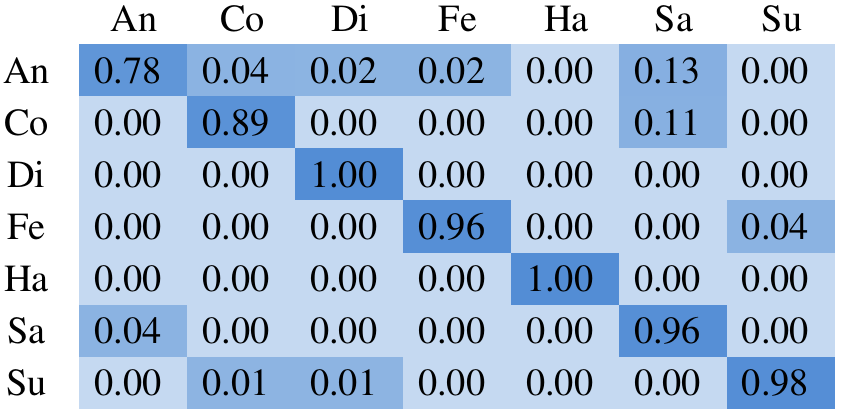}}
\subfigure[Oulu-CASIA]{
\includegraphics[height=2.1cm]{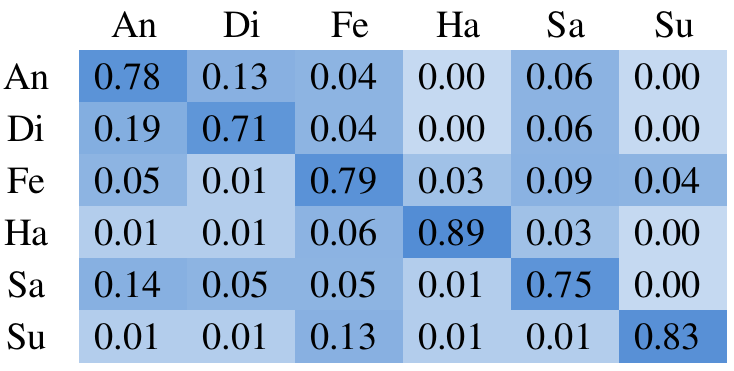}}
\subfigure[MMI]{
\includegraphics[height=2.1cm]{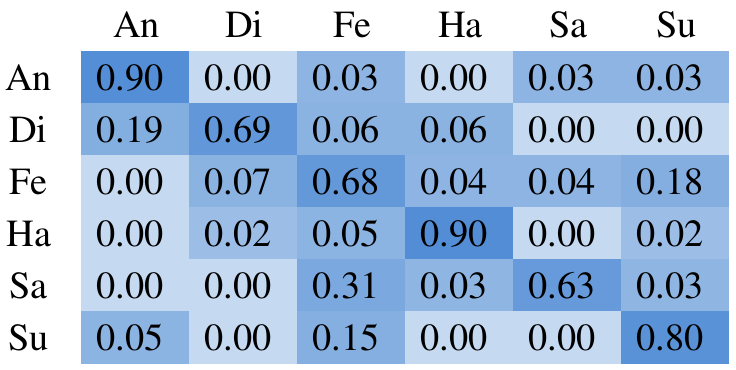}}
\subfigure[FERA]{
\includegraphics[height=2cm]{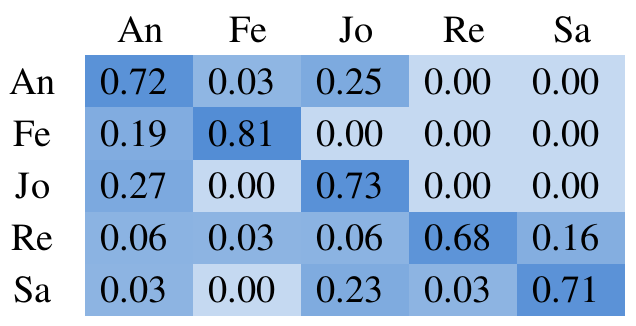}}
\caption{Confusion matrices based on ``Dis-ExpLet'' on four datasets. (a) CK+ (b) Oulu-CASIA (c) MMI (d) FERA.}
\label{fig:figConfusion}
\end{figure*}

In this section, we compare the final results with several state-of-the-art methods. Two performance metrics, i.e. the mean recognition accuracy on each category (denoted as ``mAcc'') and the overall classification accuracy (denoted as ``Acc'') are measured for comparison. The results are listed in Table~\ref{tab:tabState}. The comparisons on CK+, Oulu-CASIA, and MMI are under exactly the same protocols, and our ``ExpLet'' outperforms the existing methods significantly on both indicators (Note that, for Oulu-CASIA, ``mAcc'' is equal to ``Acc'' as the numbers of samples of all categories are the same). On FERA, by adopting cross-validation only on the training set (the same to \cite{liu2014deeply}), we compare the results with 4 most recent methods. We also review some methods in FERA challenge \cite{valstar2012meta}, in person-independent setting, our result ranks in the 2nd place, only next to the ``avatar'' based method \cite{yang2011facial} with the accuracy of 75.2\%. This may be due to that our method used fewer (6 vs. 7) subjects for training than \cite{yang2011facial}. Finally, the confusion matrices based on ``Dis-ExpLet'' on four datasets are illustrated in Figure~\ref{fig:figConfusion}. On all posed datasets, ``happy'' is always easy to be recognized, while ``fear'' and ``sad'' are more difficult and easy to be confused with each other. However, on spontanous dataset FERA, low accuracy is obtained almost on all of the categories due to the large variations in natural and different performing manners from each subject.
\begin{table*}
\caption{State-of-the-art methods on different databases. (``ExpLet*'' is the results reported in \cite{liu2014learning}.)}
\centering
\subtable[CK+]{
\begin{tabular}{ccc}
  \hline\hline
  Methods & mAcc & Acc \\
  \hline
  CLM \cite{chew2011person} & 74.4 & 82.3 \\
  AAM \cite{lucey2010extended} & 83.3 & 88.3 \\
  ITBN \cite{wang2013capturing} & 86.3 & 88.8 \\
  MCF \cite{chew2012improved} & 89.4 & -- \\
  \hline
  Fisher Vector & 89.2 & 91.7 \\
  \hline
  ExpLet* \cite{liu2014learning} & -- & 94.2 \\
  \textbf{ExpLet} & 92.8 & 94.8 \\
  \textbf{Dis-ExpLet} & \textbf{93.8} & \textbf{95.1} \\
  \hline\hline
\end{tabular}
}
\subtable[Oulu-CASIA]{
\centering
\begin{tabular}{cc}
  \hline\hline
  Methods & (m)Acc \\
  \hline
  AdaLBP(SVM) \cite{zhao2011facial} & 73.5 \\
  AdaLBP(SRC) \cite{zhao2011facial} & 76.2 \\
  LBP-TOP \cite{guo2012dynamic} & 72.8 \\
  Atlases \cite{guo2012dynamic} & 75.5 \\
  \hline
  Fisher Vector & 72.9 \\
  \hline
  ExpLet* \cite{liu2014learning} & 74.6 \\
  \textbf{ExpLet} & 76.9 \\
  \textbf{Dis-ExpLet} & \textbf{79.0} \\
  \hline\hline
\end{tabular}
}
\subtable[MMI]{
\centering
\begin{tabular}{ccc}
  \hline\hline
  Methods & mAcc & Acc \\
  \hline
  HMM \cite{wang2013capturing} & 51.5 & -- \\
  ITBN \cite{wang2013capturing} & 59.7 & 60.5 \\
  3DCNN \cite{liu2014deeply} & 50.7 & 53.2 \\
  3DCNNDAP \cite{liu2014deeply} & 62.2 & 63.4 \\
  \hline
  Fisher Vector & 68.6 & 70.7 \\
  \hline
  ExpLet* \cite{liu2014learning} & -- & 75.1 \\
  \textbf{ExpLet} & 72.4 & 73.7 \\
  \textbf{Dis-ExpLet} & \textbf{76.7} & \textbf{77.6} \\
  \hline\hline
\end{tabular}
}
\subtable[FERA]{
\centering
\begin{tabular}{ccc}
  \hline\hline
  Methods & mAcc & Acc \\
  \hline
  MSR \cite{ptucha2011manifold} & 56.6 & -- \\
  MCF \cite{chew2012improved} & 65.6 & -- \\
  3DCNN \cite{liu2014deeply} & 46.4 & 46.5 \\
  3DCNNDAP \cite{liu2014deeply} & 56.3 & 56.1 \\
  \hline
  Fisher Vector & 67.3 & 67.1 \\
  \hline
  ExpLet* \cite{liu2014learning} & -- & -- \\
  \textbf{ExpLet} & 69.5 & 69.7 \\
  \textbf{Dis-ExpLet} & \textbf{72.9} & \textbf{72.9} \\
  \hline\hline
\end{tabular}
}
\label{tab:tabState}
\end{table*}

\section{Conclusion}
In this paper, we propose a new method for dynamic facial expression recognition. By considering two critical issues of the problem, i.e. temporal alignment and semantics-aware dynamic representation, a kind of variation modeling is conducted among well-aligned spatio-temporal regions to obtain a group of expresssionlets, which serve as the mid-level representations to bridge the gap between low-level features and high-level semantics. As evaluated on four state-of-the-art facial expression benchmarks, the proposed expressionlet representation has shown its superiority over traditional methods for video based facial expression recognition. As the framework is quite general and not limited to the task of expression recognition, an interesting direction in the future is to exploit its applications in other video related vision tasks, such as action recognition and object tracking.

\section*{Acknowledgment}
\label{Ackknowledgement}

This work is partially supported by 973 Program under contract No. 2015CB351802, Natural Science Foundation of China under contracts Nos. 61222211, 61390511, 61379083, Youth Innovation Promotion Association CAS No. 2015085, and the FiDiPro program of Tekes.

\ifCLASSOPTIONcaptionsoff
  \newpage
\fi

\bibliographystyle{IEEEtran}
\balance
\bibliography{IEEEabrv,MyBib}

\end{document}